\documentclass[11pt]{article}
\usepackage{xcolor}
\usepackage{subcaption}

\usepackage{amsmath}

\definecolor{darkgreen}{rgb}{0.0,0.5,0.0}
\usepackage{booktabs}
\usepackage{titletoc}   
\usepackage{multirow} 
\usepackage[preprint]{acl}
\usepackage{amssymb} 
\usepackage{pifont}
\usepackage[T1]{fontenc} 
\usepackage{upquote} 
\usepackage{tcolorbox}
\tcbuselibrary{skins, breakable, listings} 
\usepackage{listings}
\usepackage{times}
\usepackage{latexsym}
\usepackage{booktabs}
\usepackage{graphicx}
\usepackage{cuted}  
\usepackage{multirow}
\usepackage{subcaption}

\usepackage[T1]{fontenc}

\usepackage[utf8]{inputenc}

\usepackage{microtype}

\usepackage{inconsolata}
\usepackage{amsmath}

\usepackage{graphicx}
\usepackage[table]{xcolor}

\usepackage{booktabs} 
\usepackage{tabularx} 
\usepackage{geometry} 
\usepackage{enumitem}

\usepackage{ragged2e}

\usepackage{xcolor}  

\title{RubricHub: A Comprehensive and Highly Discriminative Rubric Dataset \\ via Automated Coarse-to-Fine Generation}

\author{
    \textbf{Sunzhu Li\textsuperscript{1}},
    \textbf{Jiale Zhao\textsuperscript{1}\footnotemark[1]},
    \textbf{Miteto Wei\textsuperscript{1}\footnotemark[1]\footnotemark[2]},
    \textbf{Huimin Ren\textsuperscript{1}}, 
    \textbf{Yang Zhou\textsuperscript{3}}, \\
    \textbf{Jingwen Yang\textsuperscript{2}},
    \textbf{Shunyu Liu\textsuperscript{4}},
    \textbf{Kaike Zhang\textsuperscript{1}},
    \textbf{Wei Chen\textsuperscript{1}\footnotemark[2]}
    \\
    \textbf{\textsuperscript{1}} Li Auto Inc., China \\
        \textbf{\textsuperscript{3}} Zhejiang University
    \textbf{\textsuperscript{4}} Nanyang Technological University \\
    \textbf{\textsuperscript{2}} The Chinese University of Hong Kong, Shenzhen, China \\
    \texttt{\small \{lisunzhu, chenwei10\}@lixiang.com, vizzlin@foxmail.com}
}

\begin{document}
\maketitle

\renewcommand{\thefootnote}{\fnsymbol{footnote}}
\footnotetext[1]{\ \ Equal contribution.}
\footnotetext[2]{\ \ Corresponding author.}
\renewcommand{\thefootnote}{\arabic{footnote}}

\begin{abstract}

Reinforcement Learning with Verifiable Rewards (RLVR) has driven substantial progress in reasoning-intensive domains like mathematics. However, optimizing open-ended generation remains challenging due to the lack of ground truth. While rubric-based evaluation offers a structured proxy for verification, existing methods suffer from scalability bottlenecks and coarse criteria, resulting in a supervision ceiling effect. To address this, we propose an automated Coarse-to-Fine Rubric Generation framework. By synergizing principle-guided synthesis, multi-model aggregation, and difficulty evolution, our approach produces comprehensive and highly discriminative criteria capable of capturing the subtle nuances. Based on this framework, we introduce RubricHub, a large-scale ($\sim$110k) and multi-domain dataset. We validate its utility through a two-stage post-training pipeline comprising Rubric-based Rejection Sampling Fine-Tuning (RuFT) and Reinforcement Learning (RuRL). Experimental results demonstrate that RubricHub unlocks significant performance gains: our post-trained Qwen3-14B achieves state-of-the-art (SOTA) results on HealthBench (69.3), surpassing proprietary frontier models such as GPT-5. Our code is available at \href{https://github.com/teqkilla/RubricHub}{ this URL}.

\end{abstract}

\begin{figure}[t]
  \centering
  \includegraphics[width=1\columnwidth]{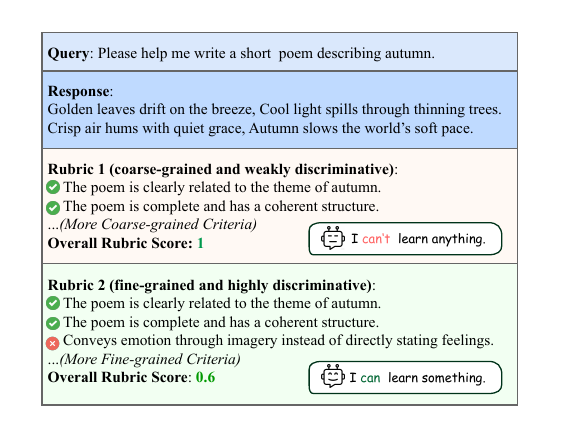}

  \caption{Motivating Example. Comparison between coarse-grained and fine-grained evaluation. Coarse rubrics (Rubric 1) result in indistinguishable high scores, whereas RubricHub (Rubric 2) utilizes highly discriminative criteria to reveal specific weaknesses, providing richer signals for alignment.}
  \label{intro:case}
\end{figure}

\section{Introduction}
Large Language Models (LLMs) are now widely deployed in real-world applications, making reliable evaluation of response quality increasingly important~\citep{zheng2023judging,chang2024survey,liang2022holistic}. In verifiable domains like mathematics and coding, Reinforcement Learning with Verifiable Rewards (RLVR) has driven substantial progress in complex reasoning, as seen in DeepSeek R1~\citep{guo2025deepseek,lambert2024tulu}. In contrast, most real-world queries are open-ended and lack ground-truth answers, leading to subjective and unstable quality judgments. Recent studies~\citep{arora2025healthbench,team2025kimi,liu2025deepseek} show that rubric-based evaluation mitigates this issue by decomposing quality into explicit, checkable criteria. By serving as a structured proxy for verification, rubrics yield interpretable assessments and more stable training signals, narrowing the gap between verifiable reasoning and open-ended generation~\citep{gunjal2025rubrics,huang2025reinforcement,zhou2025breaking}.

Despite their promise, existing rubrics face critical bottlenecks that hinder scalability.
(i) \textit{Reliance on Manual Expertise:} High-quality rubric creation demands expensive human effort, hindering its scalability.~\citep{starace2025paperbench,arora2025healthbench}. 
(ii) \textit{Narrow Domain Breadth:} Current datasets~\citep{gunjal2025rubrics} are confined to specialized domains, restricting their utility for general-purpose LLMs. 
(iii) \textit{Low Discriminability:} As illustrated in Figure~\ref{intro:case}, existing rubrics often rely on coarse, generic criteria that fail to capture subtle nuances. Consequently, they struggle to distinguish superficially plausible responses from truly high-quality ones~\citep{zhang2025chasing}, creating  \textit{ceiling effects} in supervision signals.

To overcome these bottlenecks, we propose a fully automated \textbf{Coarse-to-Fine Rubric Generation} framework. First, we synthesize candidate criteria using a response-grounded and principle-guided strategy to maintain alignment with query intent. Second, we aggregate diverse perspectives from heterogeneous models to ensure comprehensiveness, mitigating single-source biases. Crucially, to increase discriminability, we employ a difficulty evolution mechanism. Instead of stopping at generic criteria, this mechanism evolves criteria to capture the discriminative nuances of exceptional responses, ensuring the rubric remains challenging enough to guide the alignment of top-tier models. Based on this framework, we construct \textbf{RubricHub}, a large-scale ($\sim$110k), and multi-domain rubric dataset characterized by fine-grained supervision and high discriminative power.

To validate the practical utility of RubricHub, we implement a two-stage post-training pipeline: (i) Rubric-based Rejection Sampling Fine-Tuning (RuFT), where rubrics act as robust filters to curate high-quality data; and (ii) Rubric-based Reinforcement Learning (RuRL), where rubric scores serve as reward signals for policy optimization.  Experimental results demonstrate that RubricHub unlocks substantial gains. By post-training Qwen3-14B-Base, we achieve a 22.6-point lead over its official Instruct counterpart (Non-thinking) on HealthBench. Remarkably, our model even surpasses the frontier GPT-5 (69.3 vs. 67.2), despite being significantly smaller.

Our main contributions are  as follows:
\begin{itemize} [leftmargin=10pt]
\item We propose an automated Coarse-to-Fine Rubric Generation framework. It synergizes principle-guided and response-grounded synthesis, multi-model aggregation, and difficulty evolution to construct fine-grained criteria, thereby ensuring comprehensive evaluation coverage, capturing subtle quality nuances, and mitigating the supervision ceiling effect.

\item We introduce RubricHub, a large-scale ($\sim$110k) and multi-domain rubric dataset, providing  fine-grained and highly discriminative supervision for general-purpose LLMs.

\item We validate  RubricHub via a rubric-driven post-training pipeline (RuFT and RuRL), enabling Qwen3-14B to achieve SOTA performance on  HealthBench, notably outperforming  proprietary models (e.g., GPT-5).
\end{itemize}

\begin{figure*}[htbp]
  \centering
  \includegraphics[width=1.0\textwidth]{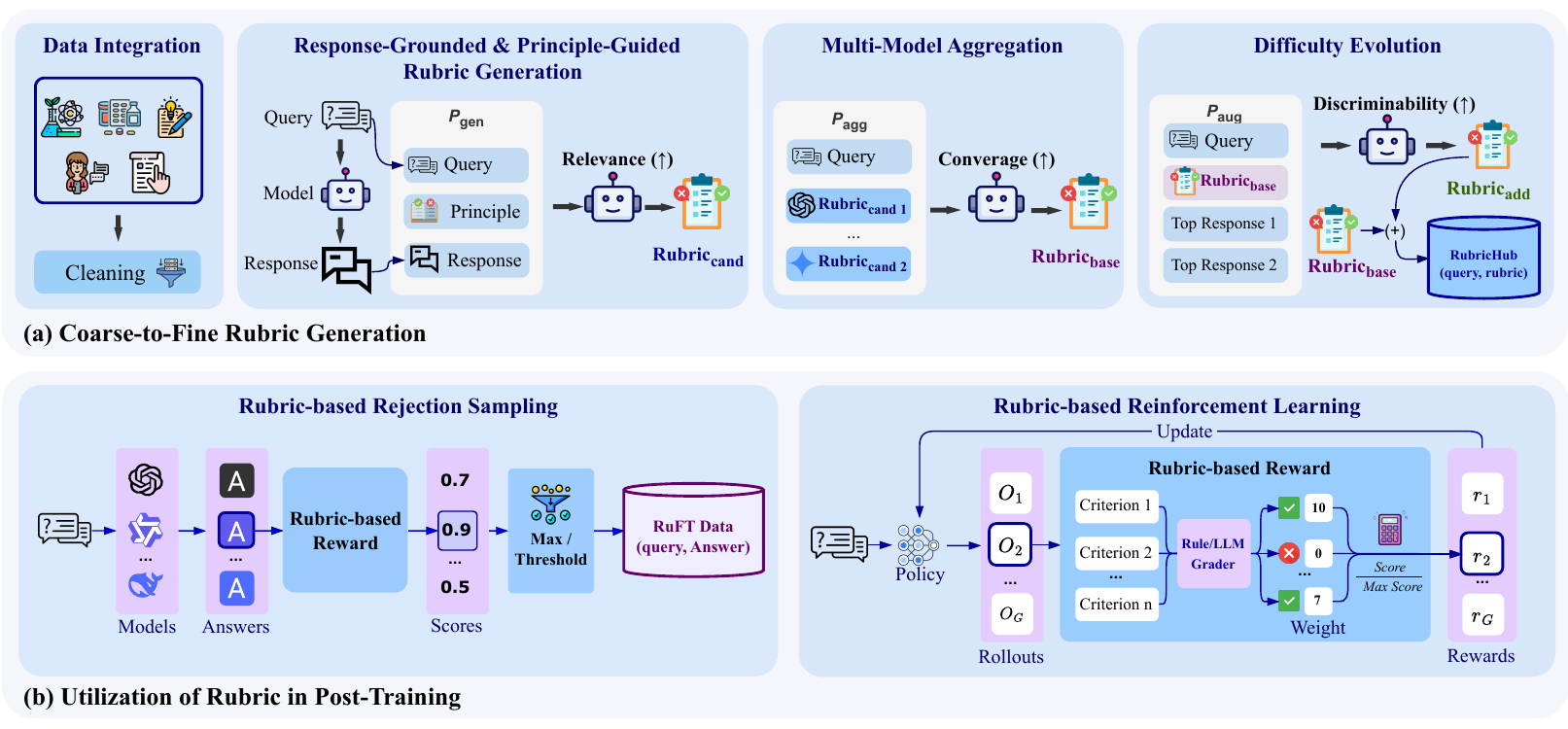}
  \caption{Overall method pipeline. (a) \textbf{Coarse-to-Fine Rubric Generation}: Candidates are synthesized via response-grounded and principle-guided strategies, then refined through aggregation and difficulty evolution into \textbf{RubricHub}. (b) \textbf{Utilization of Rubric in Post-Training }: Rubrics are applied in \textbf{RuFT} (left) for rejection sampling and in \textbf{RuRL} (right) to provide structured reward signals for policy optimization.}
  \label{main}
  \phantomsubcaption\label{main:gen}  
  \phantomsubcaption\label{main:util} 
\end{figure*}

\section{Preliminaries}
\subsection{Rubric}
Rubrics are structured scoring guides that define evaluation criteria and performance levels, widely used to assess output quality in education and model evaluation.
For each query $q$, we define a fine-grained evaluation rubric $\mathcal{R}_q$ as a set of $N_q$ weighted criteria:
\begin{equation}
\mathcal{R}_q = \{(c_i, w_i)\}_{i=1}^{N_q},
\end{equation}
where each criterion $c_{i}$ encompasses semantic requirements and grader parameters. Criteria are categorized into two types: (1) \textit{Verifiable Criteria}, representing objective constraints (e.g., format or word count) assessed via rule-based systems $\mathcal{G}_{\text{rule}}$; and (2) \textit{Semantic Criteria}, capturing qualitative attributes (e.g., reasoning depth or tone) that require LLM-based evaluators $\mathcal{G}_{\text{LLM}}$. The weight $w_{i}$ determines each criterion's importance, providing the basis for structured reward signals $r(q,o)$.

\subsection{Task Formulation}
We formulate rubric generation as a conditional task where an LLM $\mathcal{M}$ synthesizes a rubric $\mathcal{R}$ given input context $I$. By defining a prompt function $P(\cdot)$ that formats $I$ into instructions, the process is:
\begin{equation}
\mathcal{R} = \mathcal{M}\big(P(I)\big).
\end{equation}
In Section~\ref{sec:method}, we instantiate specific templates (e.g., $P_{\text{gen}}, P_{\text{agg}}$) to generate and refine rubrics through multiple stages.

\section{Method}
\label{sec:method}
In this section, we introduce our automated \textbf{Coarse-to-Fine Rubric Generation} framework. As illustrated in Figure~\ref{main}, we detail the core rubric generation pipeline, which operates in three phases: (1) \textit{Principle-Guided \& Response-Grounded Generation}, (2) \textit{Multi-Model Aggregation}, and (3) \textit{Difficulty Evolution}. Finally, we analyze the resulting dataset characteristics and detail how RubricHub is utilized for post-training.

\subsection{Coarse-to-Fine Rubric Generation}
Our core objective is to synthesize evaluation criteria that are \textit{related}, \textit{unbiased}, and \textit{highly discriminative}. Figure~\ref{main} illustrates our coarse-to-fine generation pipeline initialized with a comprehensive corpus $\mathcal{Q}$ of $\sim$110k queries, which are curated and rigorously cleaned from open-ended datasets across multiple domains. Based on this corpus, we propose a three-stage framework to synthesize and refine high-quality rubrics.

\paragraph{Stage 1: Response-Grounded \& Principle-Guided Generation.}
Generating rubrics solely from a query often leads to \textit{rubric drift}—where criteria become generic, hallucinatory, or disconnected from actual task outputs. To address this, we propose a generation strategy that is both response-grounded and principle-guided.

First, we employ response grounding by conditioning the generator $\mathcal{M}$ on a reference response $o_i$ to anchor the criteria to concrete context. Second, we enforce principle guidance by constraining the generator with a set of meta-principles $\mathbb{P}_{\text{meta}}$, encompassing:  Consistency \& Alignment; Structure \& Scope;  Clarity \& Quality; and Reasoning \& Evaluability (detailed in Appendix~\ref{appendix:meta_rubric}). Formally, using a specific generation prompt $P_{\text{gen}}$, a candidate rubric is synthesized as:
\begin{equation}
\mathcal{R}_{\text{cand}}^{(i)} = \mathcal{M}\big(P_{\text{gen}}(q, o_i, \mathbb{P}_{\text{meta}})\big).
\end{equation}
The resulting $\mathcal{R}_{\text{cand}}^{(i)}$ serves as a  context-anchored candidate, explicitly preventing the generation of generic or irrelevant criteria.

\paragraph{Stage 2: Multi-Model Aggregation.}
While Stage 1 ensures relevance, rubrics generated by a single model inherently suffer from \textit{perspective bias}. Individual models often exhibit inherent blind spots and subjective preferences, yielding narrow standards that fail to recognize valid responses with distinct presentations. To ensure comprehensiveness and objectivity, it is critical to aggregate heterogeneous viewpoints to cross-verify and mitigate these model-specific biases.

To this end, we implement multi-model aggregation. We first synthesize parallel candidate sets using heterogeneous frontier models (e.g., GPT-5.1, Gemini 3 Pro Preview) to form a unified pool $\mathcal{R}_{\text{cand}} = \bigcup_i \mathcal{R}_{\text{cand}}^{(i)}$. Subsequently, we distill this pool into a compact base rubric via an aggregation prompt $P_{\text{agg}}$, which consolidates redundant items and resolves conflicts:
\begin{equation}
\mathcal{R}_{\text{base}} = \mathcal{M}\big(P_{\text{agg}}(q, \mathcal{R}_{\text{cand}})\big).
\end{equation}
The resulting $\mathcal{R}_{\text{base}}$ serves as a comprehensive standard that explicitly eliminates single-source bias.

\paragraph{Stage 3: Difficulty Evolution.}
The base rubric $\mathcal{R}_{\text{base}}$ typically captures fundamental correctness but often lacks the granularity to distinguish between \textit{excellent} and \textit{exceptional} responses. This limitation risks score saturation, leaving top-tier models without a meaningful optimization gradient. To resolve these fine-grained quality gaps, we introduce a difficulty evolution mechanism.

Specifically, we first identify a pair of high-quality reference responses $\mathcal{A}_{\text{ref}}$, selected based on consensus high rubric scores from the initial candidate pool. We then apply an augmentation prompt $P_{\text{aug}}$ to analyze $\mathcal{A}_{\text{ref}}$, extracting discriminative nuances beyond the scope of $\mathcal{R}_{\text{base}}$ that elevate a response from \textit{excellent} to \textit{exceptional}, thereby forming a set of additive criteria $\mathcal{R}_{\text{add}}$:
\begin{equation}
    \mathcal{R}_{\text{add}} = \mathcal{M}\left(P_{\text{aug}}(q, \mathcal{R}_{\text{base}}, \mathcal{A}_{\text{ref}})\right).
\end{equation}
These criteria \textit{harden} the rubric, upgrading generic checks (e.g., ``Is the code correct?'') into rigorous standards (e.g., ``Does the code handle edge case with O(n) complexity?''). The final rubric is obtained by merging the base and evolved criteria:
\begin{equation}
    \mathcal{R}_{\text{final}} = \mathcal{R}_{\text{base}} \cup \mathcal{R}_{\text{add}}.
\end{equation}
The resulting $\mathcal{R}_{\text{final}}$ thus combines comprehensive coverage with rigorous discriminability, providing a dense and precise supervision signal for effective model optimization.

\begin{figure}[htbp]
  \centering
  \includegraphics[width=0.85\columnwidth]{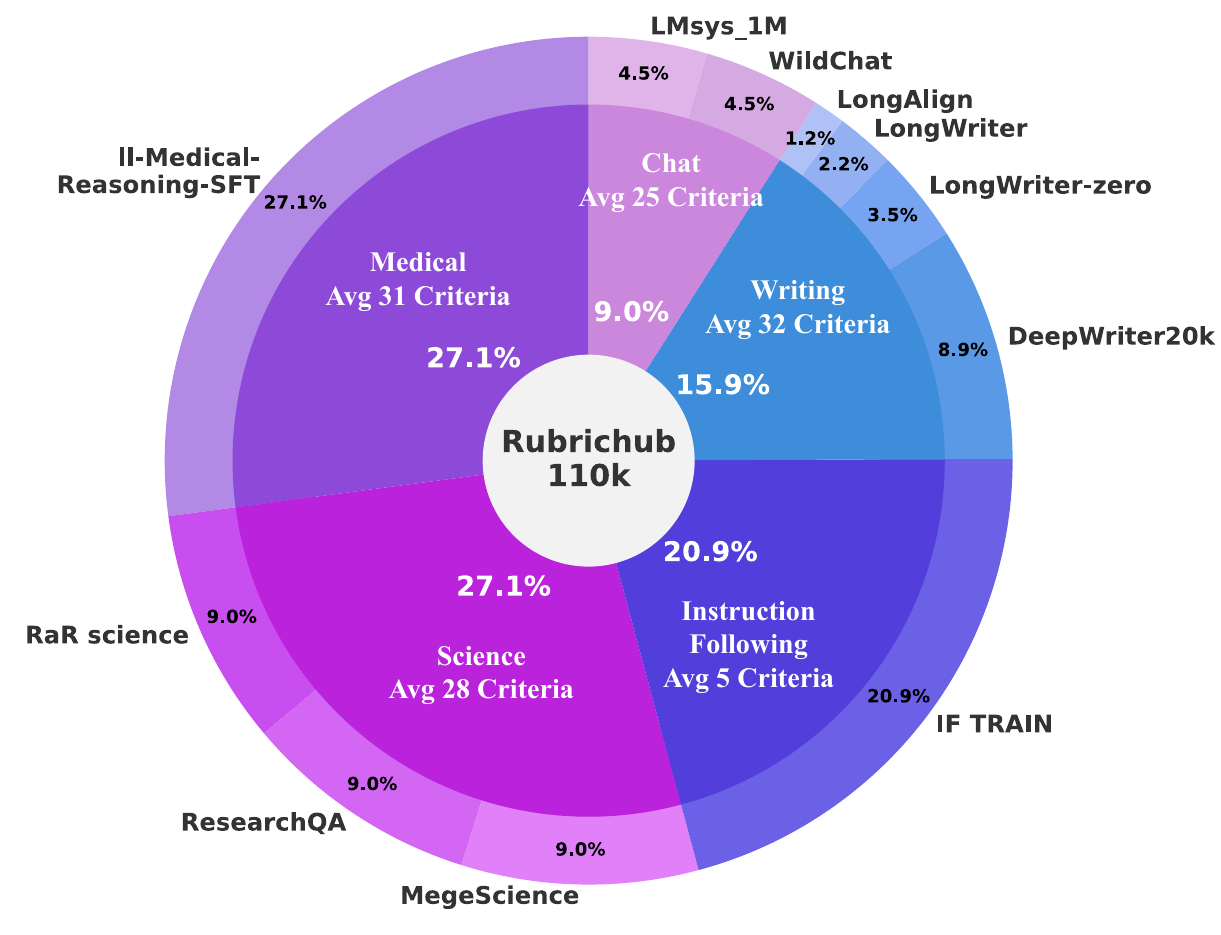}
  \caption{Pie chart showing the source distribution across five major domains.}
  \label{figure:pie}
\end{figure}

\begin{figure}[htbp]
  \centering
  \includegraphics[width=0.85\columnwidth]{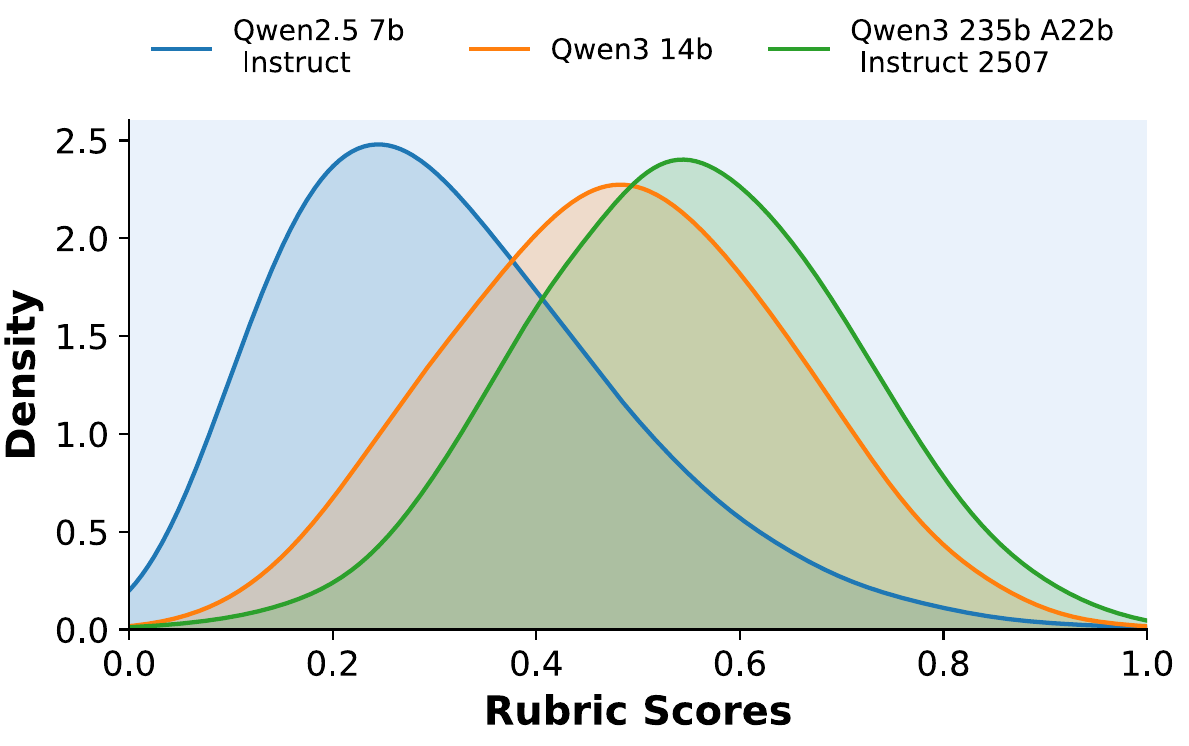}
  \caption{Score density distribution across models.}
  \label{figure:density_distribution}
\end{figure}

\subsection{Data Analysis of RubricHub}
\label{sec:data_analysis}
To construct RubricHub, we aggregated queries from five  domains:
(1) \textbf{\textit{Science}}: RaR-science~\citep{gunjal2025rubrics}, ResearchQA~\citep{yifei2025researchqa}, and MegaScience~\citep{fan2025megascience};
(2) \textbf{\textit{Instruction Following}}: IFTRAIN~\citep{pyatkin2025generalizing};
(3) \textbf{\textit{Writing}}: LongWriter~\citep{bai2024longwriter}, LongWriter-Zero~\citep{wu2025longwriter}, DeepWriting-20K~\citep{wang2025reverse}, and LongAlign~\citep{bai2024longalign};
(4) \textbf{\textit{Medical}}: II-medical~\citep{2025II-Medical-Reasoning};
(5) \textbf{\textit{Chat}}: WildChat-1M~\citep{zhao2024wildchat} and LMsys-1M~\citep{zheng2023lmsys}.

After filtering out samples with abnormal lengths or formatting errors, we sampled a final set of \textbf{$\sim$110k} question--rubric pairs. As shown in Figure~\ref{figure:pie}, RubricHub features a diverse domain composition, with Medical and Science tasks constituting the largest portions (27.1\% each), followed by Instruction Following (20.9\%) and Writing (15.9\%). The inner ring demonstrates the high density of our rubrics. For complex domains like Writing and Medical, RubricHub provides over 30 fine-grained criteria on average per query, ensuring deep and rigorous evaluation.

Crucially, the score density in Figure~\ref{figure:density_distribution} demonstrates a highly discriminative and non-saturated evaluation regime. We observe a clear distributional separation across model scales, validating the rubric's ability to distinguish varying capability levels. Moreover, even top-tier models like Qwen3-235B yield an average score of only approximately 0.6, confirming that the evolved criteria remain challenging and provide significant headroom for sustained improvement.

\subsection{Utilization of Rubrics in Post-Training}
We apply the constructed rubrics in two post-training paradigms: \textit{RuFT}, which selects high-quality data for Supervised Fine-Tuning (SFT), and \textit{RuRL}, which uses rubric scores as rewards.

\paragraph{Rubric-based Rejection Sampling Fine-Tuning.}
To ensure high-quality supervision signals, we employ a rubric-based rejection sampling strategy. For each query-rubric pair $(q, \mathcal{R}_q)$, we first prompt multiple models to generate a pool of $K$ candidate responses $\mathcal{A} = \{a_k\}_{k=1}^K$. Each response $a_k$ is independently evaluated via a scoring function $F_R$, which aggregates the weights of criteria satisfied by the response. The resulting scores are normalized to $[0,1]$:
\begin{align}
S_k &= \frac{F_R(q, \mathcal{R}_q, a_k)}{S_{\max}},
\end{align}
where $S_{\max}$ denotes the maximum achievable score for rubric $\mathcal{R}_q$. We filter out low-quality responses using a threshold $\tau$ and select the highest-scoring response:
\begin{equation}
a^+ = \arg\max_{a_k \in \mathcal{A}} \{ S_k \mid S_k > \tau \}.
\end{equation}
If no candidate exceeds $\tau$, the query is discarded. Finally, the collected high-quality pairs $\{(q, a^+)\}$ constitute the dataset used for SFT, establishing a strong initialization for subsequent alignment.

\paragraph{Rubric-based Reinforcement Learning.}
In the RL stage, the rubric defines a reward signal. For each criterion $c_i$, a unified grader $\mathcal{G}$ produces a binary score $b_i \in \{0, 1\}$:
\begin{equation}
b_i = \begin{cases} \mathcal{G}_{\text{LLM}}(q, o, c_i) & \text{for semantic criteria} \\ \mathcal{G}_{\text{rule}}(q, o, c_i) & \text{for verifiable criteria}  \end{cases}
\end{equation}
This binary formulation simplifies credit assignment and enhances training stability. The final dense reward $r(q, o)$ is calculated as the weight-normalized sum of these scores:
\begin{equation}
    r(q, o) = \frac{\sum_{i=1}^{N_q} w_i b_i}{\sum_{i=1}^{N_q} w_i},
\end{equation}
where $w_i$ represents the weight of criterion $c_i$. We optimize the policy using DAPO~\citep{yu2025dapo} under this rubric-based reward.

\section{Experiment}
\subsection{Experimental Setup}

\paragraph{Benchmarks.}

We evaluate our models on five domains spanning open-ended and closed-ended generation:
(1) \textbf{Science}: ResearchQA~\citep{yifei2025researchqa} and GPQA-Diamond~\citep{rein2024gpqa}, with accuracy as the primary metric.
(2) \textbf{Instruction-Following}: IFEval~\citep{zhou2023instruction} and IFBench~\citep{pyatkin2025generalizing}, assessing structural adherence and constraint satisfaction.
(3) \textbf{Writing}: WritingBench~\citep{wu2025writingbench} and CreateWriting-V3, emphasizing coherence, creativity, and style.
(4) \textbf{Medical}: HealthBench~\citep{arora2025healthbench} and LLMEval-Med~\citep{zhang2025llmeval}, focusing on reliability and factual accuracy.
(5) \textbf{Chat}: Arena-Hard-V2~\citep{li2024crowdsourced} and an internal dialogue survey, consistency, and multi-turn engagement.

\paragraph{Baselines.}
We compare our method against three major categories of baselines: \textbf{(1)} Proprietary models: Gemini 3.0 Pro Preview~\citep{google2025gemini}, GPT 5.1~\citep{openai2025gpt51}, GPT-4.1~\citep{openai2025gpt41} and DeepSeek V3.1~\citep{liu2024deepseek}; \textbf{(2)} Rubric-based models: Rubicon-Preview~\citep{huang2025reinforcement}, Baichuan-M2~\citep{dou2025baichuan}, and Rubrics as Reward~\citep{gunjal2025rubrics}; and \textbf{(3)} Official post-training versions of the same base model: Qwen3-4B and 14B~\citep{yang2025qwen3}.

\paragraph{Training Details.}
\label{sec:trainingsetting}
We conduct post-training on the Qwen3-4B and 14B base models. The process follows a two-stage strategy: (1) RuFT, utilizing a unified dataset of 30K high-quality instances curated via rubric-based rejection sampling for initial alignment; and (2) RuRL, where the policy is further optimized separately for each of the five domains using domain-specific datasets from RubricHub with the verl framework and the DAPO algorithm. All configuration parameters are detailed in Appendix~\ref{appendix:configuration}.

\begin{table*}[t]
\centering
\small
\setlength{\tabcolsep}{4pt}
\renewcommand{\arraystretch}{1.1}
\caption{
Broad evaluation of frontier, rubric-based, and our proposed models across five-domain benchmarks. $\dagger$ indicates results reported from official blogs, technical reports, or leaderboards. \textbf{Bold} indicates the best performance in each column within each model group. The "+" sign denotes the addition of training stages. Green and red subscripts represent the performance improvement and degradation relative to the corresponding Base model.}
\resizebox{\textwidth}{!}{
\begin{tabular}{llllllllll}
\specialrule{2pt}{0pt}{0pt}

\noalign{\vskip 2pt}

\textbf{Model} &
\multicolumn{2}{c}{\textbf{Medical}} & 
\multicolumn{2}{c}{\textbf{Instruction Following}} &
\multicolumn{2}{c}{\textbf{Writing}} &
\multicolumn{2}{c}{\textbf{Science}} & 
\textbf{Chat} \\

\cmidrule(lr){2-3}
\cmidrule(lr){4-5}
\cmidrule(lr){6-7}
\cmidrule(lr){8-9}
\cmidrule(lr){10-10}

& HealthBench & LLMEval-Med & 
IFEval & IFBench &
WritingBench & CreateWritingV3 & 
GPQA-D & ResearchQA & 
ArenaHard V2\\
\midrule

\rowcolor{gray!20} \multicolumn{10}{c}{\textbf{Proprietary Models} \rule{0pt}{6pt}} \\

Gemini3 Pro Preview  & 49.3 & 72.7 & \textbf{94.2} & \textbf{61.2} & 78.5$\dagger$ & 81.5$\dagger$ & \textbf{90.8}$\dagger$ & 77.2 & \textbf{80.8} \\

GPT 5 (high)  & \textbf{67.2}$\dagger$
  & \textbf{80.0} & - & 37.8 & \textbf{83.9}$\dagger$ & \textbf{84.0}$\dagger$ & 85.7$\dagger$ & \textbf{77.6} & 72.5 \\
  
GPT 4.1 & 47.9 & 71.2 & 87.0 & 37.2 & 69.0 & 79.0 & 50.5 & 70.8 & 49.1\\

DeepSeek V3.1 & 50.8 & 75.1 & 87.1 & 31.6 & 74.1 & 81.0 & 68.3 & 75.9 & 62.4 \\
\midrule

\rowcolor{gray!20} \multicolumn{10}{c}{\textbf{Rubric-based Models} \rule{0pt}{6pt}} \\

DR-Tulu-8B & 50.2$\dagger$ & 51.9 & 30.1  & 26.5  & 37.0  & 46.3  & 58.1  & 74.3$\dagger$  &29.6  \\

Rubicon-preview-30B-A3B & 50.4 & 73.3 & 82.9 & 33.6 & 72.8 & 66.8 & 63.6 & 74.9 & 45.0 \\

Baichuan-M2-32B & \textbf{58.8} & \textbf{79.3} & \textbf{83.6} & \textbf{38.8} & \textbf{79.2} & \textbf{72.2} & \textbf{66.2} & \textbf{75.3} & \textbf{45.8} \\

\midrule

\rowcolor{gray!20} \multicolumn{10}{c}{\textbf{Ours} \rule{0pt}{6pt}} \\

Qwen3-4B (Non-thinking) & 37.3 & 61.5 & 80.6 & 23.1  & 55.9 & 40.6 & 45.5 & 65.0 & 20.6 \\
Qwen3-4B-Base  & 0.1 & 28.3 & 34.9 & 13.5 & 34.8 & 25.4  & 36.2 & 40.9 & 0.1  \\

\quad + RuFT   & $39.4_{\color{green!50!black}{+39.3}}$  & $56.2_{\color{green!50!black}{+27.9}}$  &  $72.6_{\color{green!50!black}{+37.7}}$ & $20.4_{\color{green!50!black}{+6.9}}$ & $67.6_{\color{green!50!black}{+32.8}}$ & $39.6_{\color{green!50!black}{+14.2}}$ & $34.7_{\color{red!50!black}{-1.5}}$ & $70.1_{\color{green!50!black}{+29.2}}$ & $11.2_{\color{green!50!black}{+11.1}}$ \\

\quad + RuRL  & $60.3_{\color{green!50!black}{+46.4}}$ & $69.1_{\color{green!50!black}{+40.8}}$ & $79.1_{\color{green!50!black}{+44.2}}$ & $29.3_{\color{green!50!black}{+15.8}}$ & $71.2_{\color{green!50!black}{+36.4}}$ & $40.0_{\color{green!50!black}{+14.6}}$ & $47.2_{\color{green!50!black}{+11.0}}$ & $82.7_{\color{green!50!black}{+41.8}}$  & $29.9_{\color{green!50!black}{+29.8}}$  \\

\quad + RuFT $\rightarrow$ RuRL & 
$\textbf{65.1}_{\color{green!50!black}{+65.0}}$ & $\textbf{82.9}_{\color{green!50!black}{+54.6}}$ & $\textbf{91.4}_{\color{green!50!black}{+56.5}}$ & $\textbf{45.9}_{\color{green!50!black}{+32.4}}$ & $\textbf{74.1}_{\color{green!50!black}{+39.3}}$ & $\textbf{43.9}_{\color{green!50!black}{+18.5}}$ & $\textbf{48.5}_{\color{green!50!black}{+12.3}}$ & $\textbf{83.5}_{\color{green!50!black}{+42.6}}$ & $\textbf{54.5}_{\color{green!50!black}{+54.4}}$

\\

\cmidrule(lr){1-10}

Qwen3-14B (Non-thinking) & 46.7 & 70.2 & 85.6 & 28.2 & 63.6 & 64.6 & 51.1 & 65.9 & 21.0 \\
Qwen3-14B-Base & 22.8 & 50.3 & 49.5 & 16.4 & 44.9 & 36.0 & 38.8 & 54.9 & 5.2 \\

\quad + RuFT  & $44.4_{\color{green!50!black}{+21.6}}$ & $67.3_{\color{green!50!black}{+17.0}}$ & $80.0_{\color{green!50!black}{+30.5}}$ & $21.4_{\color{green!50!black}{+5.0}}$ & $72.3_{\color{green!50!black}{+27.4}}$ & $66.9_{\color{green!50!black}{+30.9}}$ & $45.8_{\color{green!50!black}{+7.0}}$ & $74.2_{\color{green!50!black}{+19.3}}$ & $34.9_{\color{green!50!black}{+29.7}}$ \\

\quad + RuRL & $66.2_{\color{green!50!black}{+43.4}}$ & $79.5_{\color{green!50!black}{+29.2}}$ & $85.0_{\color{green!50!black}{+35.5}}$ & $37.1_{\color{green!50!black}{+20.7}}$ & $76.3_{\color{green!50!black}{+31.4}}$ & $62.9_{\color{green!50!black}{+26.9}}$ & $58.4_{\color{green!50!black}{+19.6}}$ & $85.5_{\color{green!50!black}{+30.6}}$ & $65.6_{\color{green!50!black}{+60.4}}$ \\

\quad + RuFT $\rightarrow$ RuRL & $\textbf{69.3}_{\color{green!50!black}{+46.5}}$ & $\textbf{83.2}_{\color{green!50!black}{+32.9}}$ & $\textbf{92.6}_{\color{green!50!black}{+43.1}}$ & $\textbf{51.4}_{\color{green!50!black}{+35.0}}$ & $\textbf{79.4}_{\color{green!50!black}{+34.5}}$ & $\textbf{70.4}_{\color{green!50!black}{+34.4}}$ & $\textbf{58.5}_{\color{green!50!black}{+19.7}}$ & $\textbf{86.2}_{\color{green!50!black}{+31.3}}$ & $\textbf{74.4}_{\color{green!50!black}{+69.2}}$ \\

\specialrule{2pt}{0pt}{0pt}
\end{tabular}}
\label{tab:rubric-benchmarks}
\end{table*}

\begin{figure*}[htbp]
  \centering
  \includegraphics[width=1.0\textwidth]{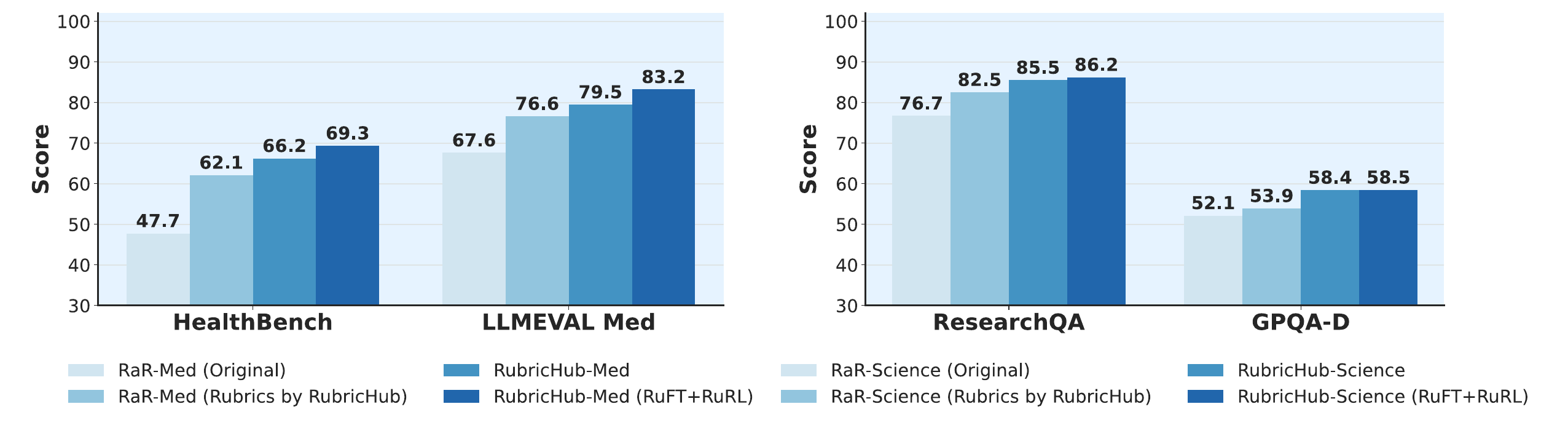}
  \caption{Performance comparison using RaR and RubricHub in Medical (left) and Science (right) domains on Qwen3-14B-Base. \textit{RaR (original)}: original RaR dataset. \textit{RaR (Rubrics by RubricHub)}: RaR questions with Rubrics regenerated by our pipeline.}
  \label{expr}
\end{figure*}

\subsection{Main Results}

\paragraph{Comparison of Post-Training Schemes.} 
Results across Qwen3-4B and 14B reveal a consistent performance hierarchy across all  domains: \textit{Base} < \textit{RuFT} < \textit{RuRL} < \textit{RuFT$\rightarrow$RuRL}. Notably, the pipeline achieves its largest gain in general chat capabilities: on ArenaHard V2, the Qwen3-14B score surges from 5.2 (Base) to 74.4, demonstrating the method's effectiveness in unlocking latent model potential. This validates our multi-stage strategy: RuFT provides a supervised \textit{cold start} for task alignment, establishing a foundation that enables RuRL to further maximize performance.

\paragraph{Comparison with Frontier and Rubric-Based Models.}
Our proposed models not only outperform rubric-based baselines but also achieve competitive results against top-tier proprietary models. Compared to the larger Baichuan-M2-32B, our Qwen3-14B prevails in 4 out of 5 domains (Medical, Instruction Following, Chat, and Science), highlighting the superior quality of our alignment recipe. Against proprietary giants, it achieves competitive results on general benchmarks, surpassing GPT-4.1 and DeepSeek V3.1 on IFEval (92.6) and ArenaHard V2 (74.4). Most notably, in the medical domain, it achieves SOTA performance with a score of 69.3 on HealthBench, outperforming even the frontier GPT-5 (67.2).

\begin{table}[t]
\centering
\small
\setlength{\tabcolsep}{2pt}
\renewcommand{\arraystretch}{1.05}
\caption{Impact of different grader models on medical performance. $\ddag$ denotes the Instruct-2507 version.}
\label{tab:Grader}
\resizebox{.48\textwidth}{!}{
\begin{tabular}{lccc}

\toprule
\textbf{Grader} & \textbf{HealthBench}   & \textbf{LLMEval-Med} \\
\midrule

Qwen2.5-7B-Instruct  & 60.3  & 71.8\\
Qwen3-30B-A3B$\ddag$  & 62.3   &  71.8\\
Qwen3-235B-A22B$\ddag$  &  66.4  & 77.7 \\
\textbf{gpt-oss-120B Auto (Used)}  & 66.2   & 79.5 \\

\bottomrule

\end{tabular}
}

\end{table}

\paragraph{Comparison with Open-Source Rubric Data.} 
Given the scarcity of publicly available rubric datasets, we benchmark our method against the representative RaR rubrics. As illustrated in Figure~\ref{expr}, our pipeline-generated rubrics significantly improve supervisory quality compared to the original RaR rubrics. We observe a dramatic improvement on HealthBench (47.7 to 62.1) and a steady gain on ResearchQA (76.7 to 82.5) when switching to rubrics generated by RubricHub. Moreover, employing the full RubricHub dataset yields further improvements (3rd bar). Finally, applying the full RuFT$\rightarrow$RuRL pipeline maximizes performance (4th bar), achieving the best results across these experimental settings.

\subsection{Analysis}

\begin{figure}[t]
    \centering
    \includegraphics[width=\columnwidth]{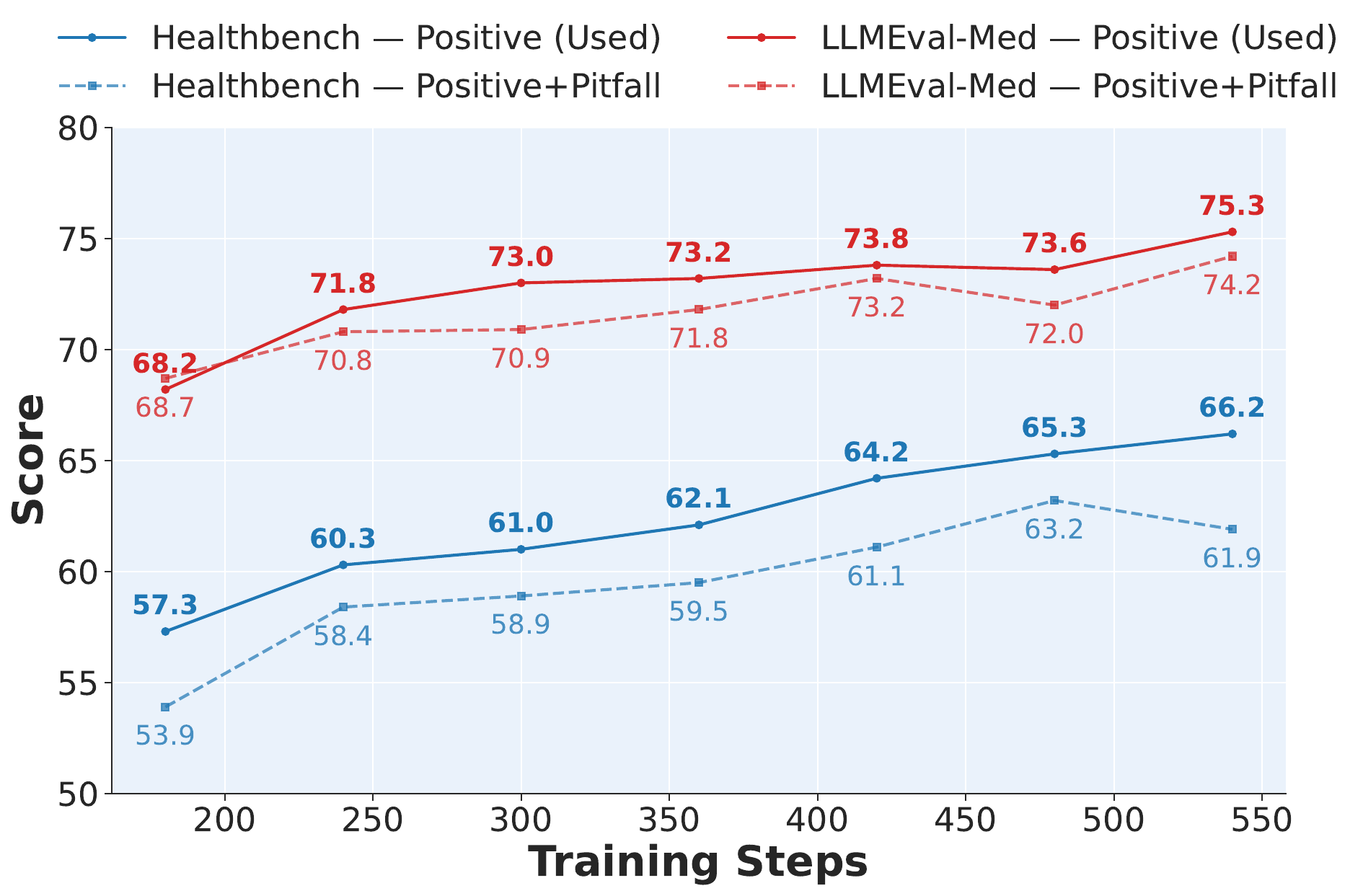}
    \caption{ Effect of criteria composition on RL performance (Qwen3-14B-Base). Training with only positively weighted criteria (Positive, ours) consistently outperforms the inclusion of negative penalties (Positive + Pitfall) across both benchmarks.}
    \label{fig:pitfall}
\end{figure}

\paragraph{Sensitivity Analysis.}

To assess the impact of rubric \textbf{criteria types} and \textbf{grader models}, we conducted a sensitivity analysis on medical benchmarks using Qwen3-14B-Base (RuRL). Regarding \textit{criteria}, Figure~\ref{fig:pitfall} shows positive-only weights consistently outperform those with negative penalties, achieving higher scores on HealthBench (66.2 vs. 63.2) and LLMEval-Med (75.3 vs. 74.2). We attribute this to the grader’s low accuracy on negative criteria~\citep{arora2025healthbench}, which hinders optimization; thus, we adopt positive-only formulation.  
For \textit{grader} models (Table~\ref{tab:Grader}), Qwen2.5-7B and Qwen3-30B-A3B are weak. Qwen3-235B-A22B possesses the largest parameter scale, and its inference latency is several times higher than other candidates, making it prohibitively slow for large-scale iterations. After balancing effectiveness and speed, we select gpt-oss-120B as our grader.

\begin{figure}[t]
    \centering
    \includegraphics[width=\columnwidth]{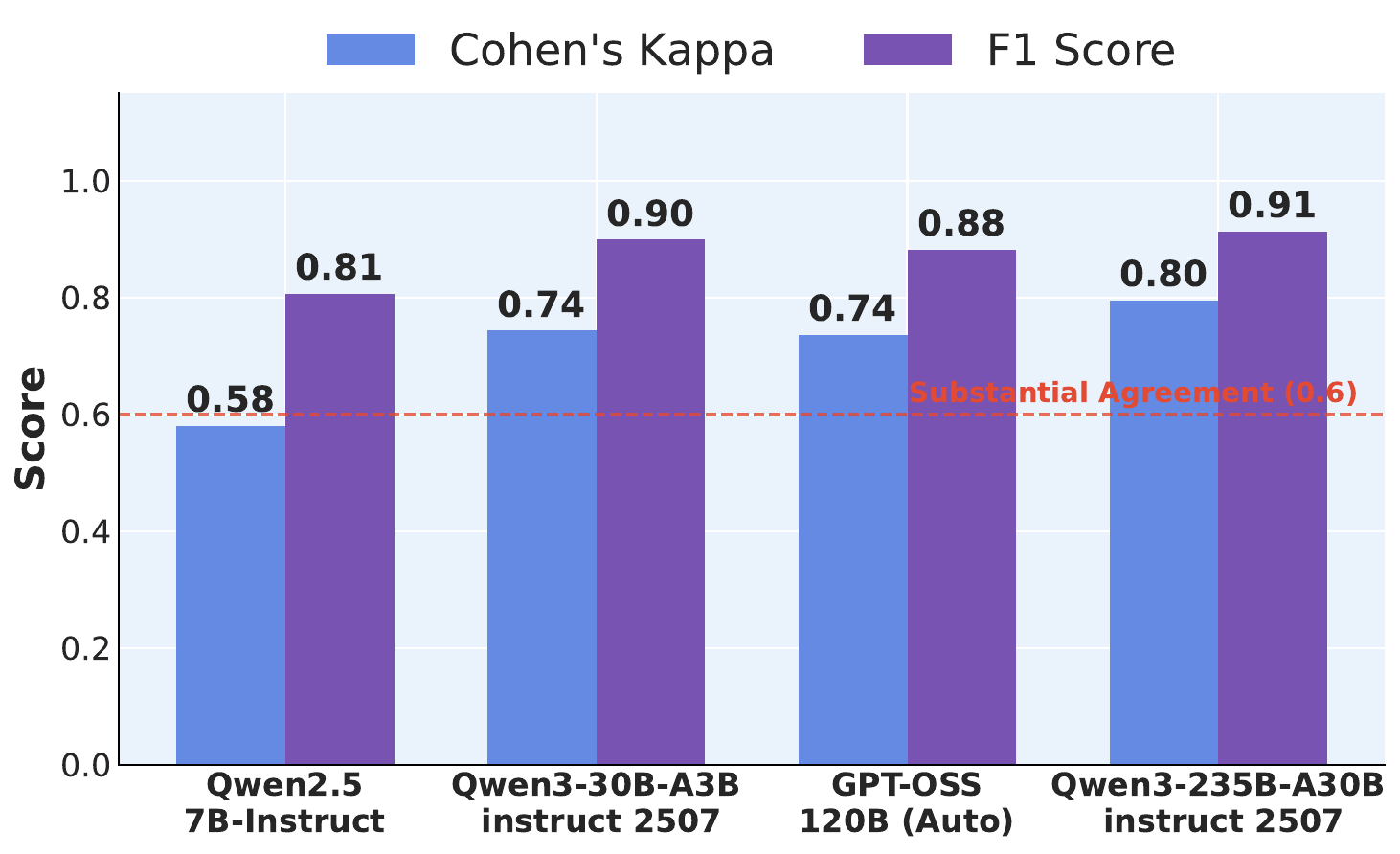}
    \caption{Agreement between Human and LLM evaluations. Blue bars: \textbf{Cohen's Kappa }for inter-rater reliability. Purple bars: \textbf{F1 Score} treats human scores as ground truth.  Red dashed line (0.6): threshold for substantial agreement.}
    \label{fig:llm_human}
\end{figure}

\paragraph{Agreement Between Human and LLM.}

As illustrated in Figure~\ref{fig:llm_human}, we evaluated rubric robustness by comparing human judgments with LLMs ranging from 7B to 235B across 940 criteria. Results reveal a scale-dependent improvement from 7B to 30B: the 7B baseline shows moderate agreement (F1 Score: 0.81, $\kappa$: 0.58), while the 30B model achieves higher consistency (F1 Score: 0.90, $\kappa$: 0.74), indicating a capability threshold for reliable evaluation. Beyond this point, performance saturates, with only marginal variance among the 30B, 120B, and 235B models ($\kappa$: 0.74–0.80). This convergence suggests that the rubric generalizes well across high-capacity models and is  insensitive to further increases in model scale.

\begin{figure}[t]
    \centering
    \includegraphics[width=\columnwidth]{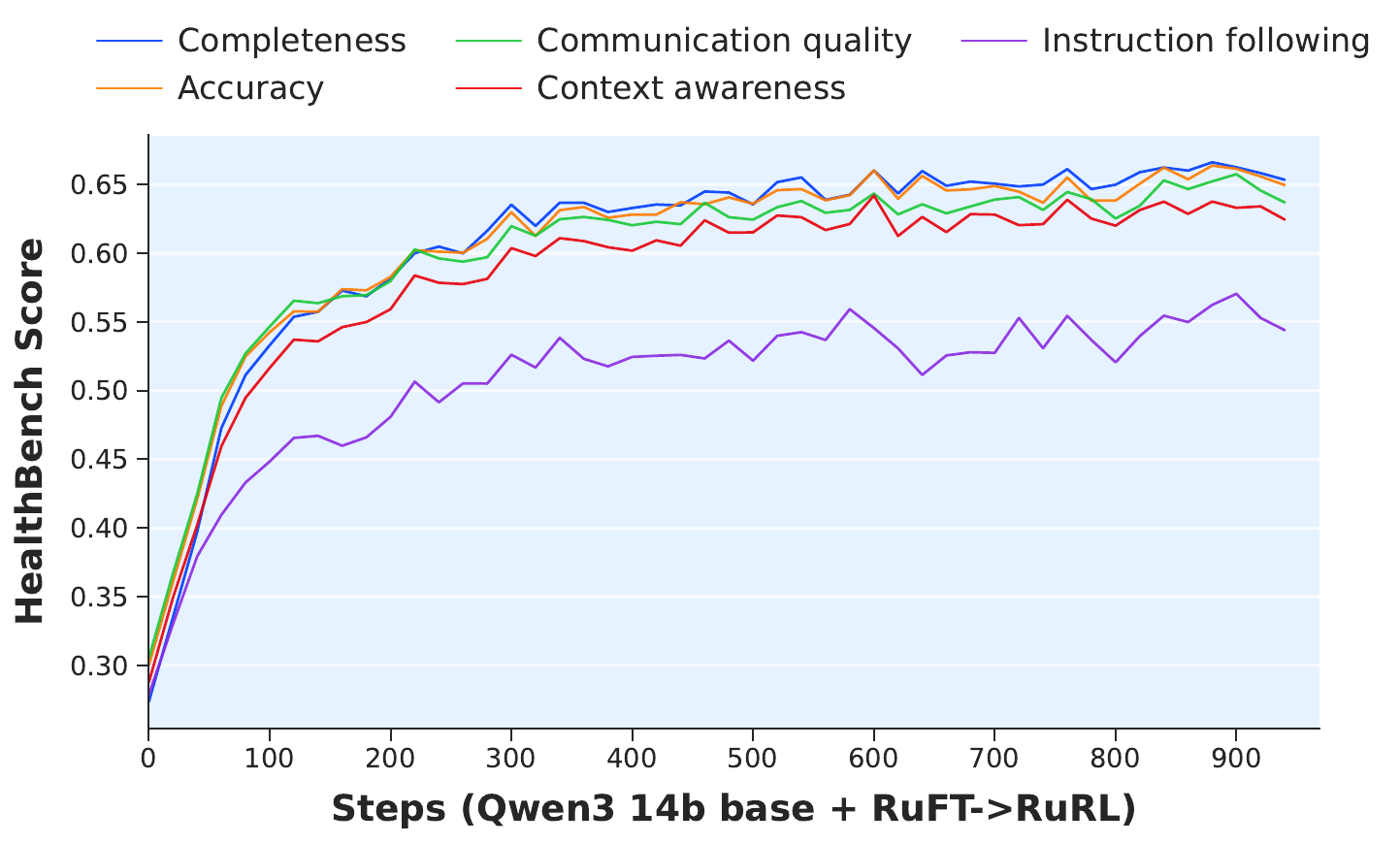}
    \caption{Training dynamics analysis on the HealthBench test set, with five colored lines corresponding to the rubric dimensions.}
    \label{fig:training_analysis}
\end{figure}

\paragraph{Training Dynamics Analysis.}
Figure~\ref{fig:training_analysis} shows the model's performance trajectory on HealthBench during training, yielding two key observations. First, the improvement is \textit{steady}. Scores rise rapidly and converge, validating our \textit{RubricHub} (and RuRL) strategy. Second, the growth is \textit{balanced}. The synchronized rise in metrics like Accuracy, Completeness, and Communication Quality indicates holistic capability enhancement rather than over-optimization for a single dimension.

\begin{table}[t]
    \centering
    \caption{Ablation study of the Coarse-to-Fine Rubric Generation Pipeline. The marker (+) indicates the cumulative addition of components. \textit{Naive Rubric Gen.}: Direct generation via a single model (GPT-5.1); \textit{PG 
    \& RG}: Adds Principle-Guided and Response-Grounded constraints; \textit{Multi-Model Agg.}: Aggregates candidates from multiple models; \textit{Difficulty Evolution (Full)}: Incorporates difficulty evolution to complete the pipeline.
    }
    
    \label{tab:Ablation_Pipeline}
    \resizebox{\linewidth}{!}{
        \begin{tabular}{lcc}
            \toprule
            \textbf{Method Setting} & \textbf{HealthBench} & \textbf{LLMEval-Med} \\
            \midrule
            \textit{Naive Rubric Gen.}   & 60.9  & 71.7  \\
            + \textit{PG \& RG}    & 63.8 & 74.1 \\
            + \textit{Multi-Model  Agg. }  & 65.0 & 75.6 \\
            + \textit{\textbf{Difficulty Evolution  (Full)}}  & 66.2 & 79.5 \\
            \bottomrule
        \end{tabular}
    }
\end{table}

\begin{figure}[t]
    \centering
    \includegraphics[width=\columnwidth]{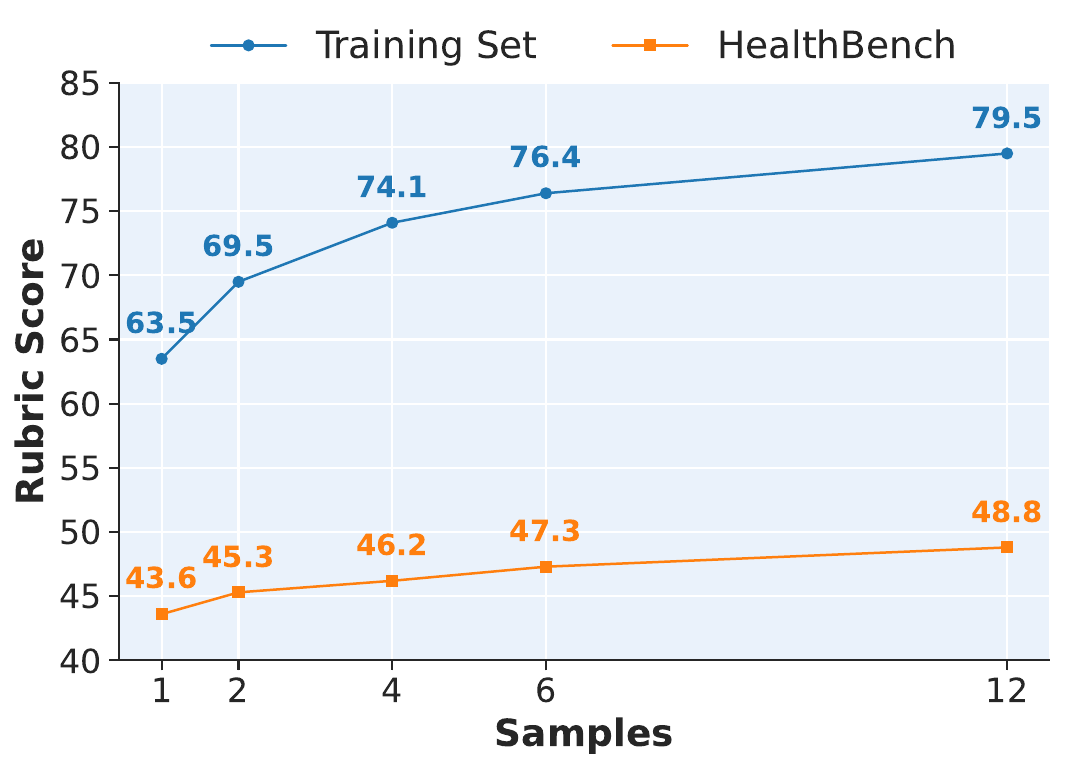}
    \caption{Ablation of Rubrics-based Rejection Sampling Fine-Tuning. \textbf{Samples} denotes the number of  answers per question. \textbf{Rubric Score}: On the \textit{Training Set}, we first select the highest-scoring sampled response for each question and then average these scores; \textit{HealthBench} scores follow the official evaluation protocol. }
    \label{fig:sft}
\end{figure}

\subsection{Ablation Study}

\paragraph{Ablation Study of Coarse-to-Fine Rubric Generation.} 
As shown in Table~\ref{tab:Ablation_Pipeline}, we conduct an incremental ablation study to validate our framework. Compared to the \textit{Naive Rubric Gen.} baseline, adding Principle-Guided and Response-Grounded constraints (\textit{+ PG \& RG}) yields a notable improvement (\textit{e.g.}, +2.9 on HealthBench and +2.4 on LLMEval-Med), demonstrating the importance of constrained generation. The \textit{Multi-Model Agg.} component further enhances performance by reducing single-model bias. Finally, incorporating \textit{Difficulty Evolution} completes the framework, resulting in the most significant gains on LLMEval-Med (reaching 79.5). The strictly monotonic improvements across both benchmarks confirm the additive value of each component in our Coarse-to-Fine framework.

\paragraph{Ablation Study of Rubric-based Rejection Sampling Fine-Tuning.}

Figure \ref{fig:sft} shows an ablation of Rubric-based rejection sampling across varying sample sizes ($n$). Increasing candidates from 1 to 12 raises the average maximum Training Set score from 63.45 to 79.51, elevating the quality upper bound. Models trained on this refined data show steady improvement on HealthBench, rising from 43.61 to 48.81. These results show that increasing candidate quantity with Rubric-based filtering enhances final output quality.

\section{Related Works}

\subsection{LLM-as-a-Judge and Rubric Evaluation}

As LLM outputs become increasingly open-ended, evaluating response quality has become a central challenge. The \emph{LLM-as-a-Judge} paradigm addresses this by using LLMs to assess model-generated responses~\citep{zheng2023judging}. However, directly assigning coarse-grained scores (e.g., Likert ratings) is often unstable and biased~\citep{wang2024large}. To improve reliability, recent work adopts \emph{rubric-based evaluation}, which decomposes quality into interpretable criteria~\citep{wang2024large,gunjal2025rubrics}. Several benchmarks across domains leverage expert-authored rubrics to enable more structured and consistent evaluation of complex responses~\citep{arora2025healthbench,starace2025paperbench,wang2025profbench}.

\subsection{Rubric Data Automatic Generation}

To enable scalable rubric-style supervision, recent work has explored automatic rubric construction beyond expert-designed criteria~\citep{arora2025healthbench,wang2025profbench}. Existing methods broadly fall into three categories:
\textit{(i) LLM-synthesized rubrics}, which prompt LLMs to generate evaluation criteria for a given task~\citep{gunjal2025rubrics,huang2025reinforcement};
\textit{(ii) rubrics mined from human-authored documents}, which extract and structure evaluation dimensions from high-quality resources such as academic surveys or web content~\citep{yifei2025researchqa,anonymous2025qurl};
and \textit{(iii) rubrics induced from preference data}, which infer reusable evaluation dimensions from pairwise comparison signals~\citep{liu2025openrubrics,wang2025autorule}. Our work builds on this line by further improving the scalability and quality of automatically generated rubrics.

\section{Conclusion}

To address the lack of  ground truth in open-ended tasks, this work introduces an automated Coarse-to-Fine rubric generation framework and establishes RubricHub—a large-scale ($\sim$110k) and multi-domain rubric dataset characterized by high discriminability. By synergizing principle-guided and response-grounded synthesis, multi-model aggregation, and  difficulty evolution, our approach constructs comprehensive and fine-grained criteria that cover diverse quality dimensions while resolving subtle differences among high-performing model outputs, effectively alleviating the supervision ceiling effect that limits existing rubric-based methods. By leveraging these rubrics to drive Rejection Sampling Fine-Tuning (RuFT) and Reinforcement Learning (RuRL), a Qwen3-14B model achieves significant performance gains, surpassing proprietary giants like GPT-5 on benchmarks such as HealthBench. This work demonstrates the efficacy of fine-grained rubrics as a scalable, automated solution for model alignment.

\section{Limitations}
Despite the advancements of RubricHub, several limitations remain:

\textbf{Domain Scope:} Although RubricHub includes certain scientific reasoning tasks (e.g., GPQA-Diamond), it primarily addresses non-verifiable domains and lacks systematic coverage of purely verifiable tasks such as complex mathematics and competitive coding. Furthermore, long-horizon agentic tasks requiring multi-step planning remain unexplored.

\textbf{Grader Reliability and Capacity:} Incorporating Pitfalls introduces significant noise that degrades RL performance. This instability is fundamentally exacerbated by model scale; compact models fall below the capability threshold for reliable evaluation even when restricted to positive criteria. This necessitates a reliance on costly large-scale graders and highlights the need for specialized, high-precision compact grader architectures.

\textbf{Efficiency:} Rubric-driven training, particularly during the RuRL stage, involves substantial computational overhead and inference latency. While parallel grader deployment partially mitigates these issues , further architectural optimizations—such as hybrid serial-parallel scoring—are required for efficient large-scale iterations.

\bibliography{custom}

\clearpage
\appendix
\section*{Appendix}
\section*{Table of Contents}
\hypersetup{
  linkcolor=darkblue
}
\startcontents[sections]
\printcontents[sections]{l}{1}{\setcounter{tocdepth}{2}}

\newpage

\section{High Quality Rubric Principle}
\label{appendix:meta_rubric}

\begin{table*}[htbp]
\small
\centering
\caption{High-quality Rubric dimensions and criteria. These dimensions evaluate the quality of other rubrics by assessing clarity, coherence, structure, and logical alignment of their criteria with the intended task objectives.}
\renewcommand{\arraystretch}{1.1} 
\begin{tabular}{p{3.5cm} p{11.5cm}} 
\toprule
\textbf{Dimensions} & \textbf{Criterion Description} \\
\midrule
\multicolumn{2}{l}{\textit{\textbf{Consistency and Alignment}}} \\
\noalign{\vskip 2pt}
Consistency & The rubric should yield highly consistent scores when used by three or more graders. \\
\addlinespace
Stability & The rubric should be consistent for the same grader on the same query ($\geq 3$ times). \\
\addlinespace
Alignment & Each criterion should be an Explicit, Implicit, or Pitfall item relevant to the query. \\

\midrule
\multicolumn{2}{l}{\textit{\textbf{Structure and Scope}}} \\
\noalign{\vskip 2pt}
Coverage & The rubric should cover all explicit instructions and implicit requirements. \\
\addlinespace
Criteria Num. & The rubric should contain between 3 and 25 criteria. \\
\addlinespace
Independence & Each criterion should not strongly depend on, contradict, or overlap with others. \\
\addlinespace
Atomicity & Each criterion should assess one independent dimension only. \\

\midrule
\multicolumn{2}{l}{\textit{\textbf{Clarity and Quality}}} \\
\noalign{\vskip 2pt}
Clarity & Each criterion should be explicit and unambiguous, avoiding the use of vague terms. \\
\addlinespace
Conciseness & Each criterion should be 5–40 characters or 1–4 sentences long. \\
\addlinespace
Lang Consist. & Each criterion should use the same language as the question. \\

\midrule
\multicolumn{2}{l}{\textit{\textbf{Reasoning and Evaluability}}} \\
\noalign{\vskip 2pt}
Distinguishability & The rubric should distinguish response quality and model performance. \\
\addlinespace
Weight Rationality & Each criterion should have a weight ranging from -10 to 10. \\
\addlinespace
Verifiability & Each criterion should be verifiable through observable evidence. \\

\bottomrule
\end{tabular}
\label{tab:meta-rubric}
\end{table*}
\newpage

\section{Detailed Training Settings}
\label{appendix:configuration}

\begin{table}[!h]
\vspace{-0.1cm}
\caption{RL training configuration.}
\label{tab:training_config}
\centering
\small
\begingroup
\setlength{\tabcolsep}{2pt}
\renewcommand{\arraystretch}{1.02}
\begin{tabularx}{\columnwidth}{@{}l X@{}}
\toprule
\textbf{Category} & \textbf{Configuration} \\
\midrule
\multirow{3}{*}{DAPO} &
RL Algorithm: DAPO\\
& Clip: $\epsilon_{\text{low}}=0.2$, $\epsilon_{\text{high}}=0.28$, $c=10.0$ \\
& Overlong Buffer=4096, Penalty=0.5 \\
\midrule
\multirow{1}{*}{Backbone} &
Model: Qwen3-14B-Base \\
\midrule
\multirow{5}{*}{Sampling} &
Train Temperature: 1.0 \\
& Train Top-P: 1.0, Top-K: -1 \\
& Rollout Samples per Prompt: 8 \\
& Max Prompt Length: 4096 \\
& Max Response Length: 8192 \\
\midrule
\multirow{7}{*}{Training} &
Optimizer: AdamW \\
& Learning Rate: $1\times 10^{-6}$ (constant) \\
& Warmup Steps: 10 \\
& Weight Decay: 0.1 \\
& Training Batch Size: 64 \\
& Mini Batch Size: 32 \\
& KL Loss Coefficient: 0 \\
& Total Training Steps: 500 \\
\midrule
\multirow{1}{*}{Hardware} &
GPUs: 8 $\times$ H200 \\
\bottomrule
\end{tabularx}
\endgroup
\vspace{-0.2cm}
\end{table}

We conduct post-training on two base models, Qwen3-14B and Qwen3-4B.

For \textbf{RuFT}, we construct a dataset of 30K instances via rubric-based rejection sampling (threshold $\tau=0.6$). Specifically, for randomly sampled prompts, we generate six candidate responses using GPT-5.1 and retain the highest-scoring candidate that satisfies the quality threshold. This curated dataset serves as the initialization for RuRL and is used for mixed training via LlamaFactory~\citep{zheng2024llamafactory}. We train for 3 epochs with a batch size of 64 and a cutoff length of 20480, using AdamW with a learning rate of $1\times10^{-5}$, cosine decay to $1\times10^{-6}$, and 20 warmup steps.

For \textbf{RuRL}, we train on the full RubricHub dataset ($\sim$110K instances) using the verl framework~\citep{sheng2024hybridflow}. To preserve domain-specific characteristics, RL is performed separately for each domain up to 5 epochs with DAPO. We use a batch size of 64 (mini-batch 32) and AdamW with a learning rate of $1\times10^{-6}$. KL regularization is removed by disabling KL in both the reward and loss. For each prompt, 8 rollouts are sampled with temperature 1.0 and no Top-p/Top-k sampling. The maximum prompt and response lengths are 4096 and 8192, respectively. To discourage overly long outputs, Overlong Reward Shaping is applied with a soft buffer (buffer length 4096, penalty factor 0.5). Clipping bounds are set to $\varepsilon_{\text{low}}=0.2$ and $\varepsilon_{\text{high}}=0.28$. Key hyperparameters are summarized in Table~\ref{tab:training_config}.

\section{Additional Related Work}
\label{appendix:Related_Work}
\subsection{RL for LLMs}
Early alignment method for LLMs mainly relied on human preference feedback. Representative methods such as RLHF and DPO use human-labeled comparisons of response quality to train reward models and guide policy optimization~\citep{ouyang2022training,rafailov2023direct}.
On the other hand, reinforcement learning with verifiable rewards (RLVR) has emerged, using objectively checkable outcomes of a task (e.g., code unit tests, whether a math solution is correct) as reward signals~\citep{guo2025deepseek,lambert2024tulu,yu2025dapo}.
However, RLVR is constrained by the requirement that tasks have a clear ground truth, making it difficult to apply directly to settings without a ground truth.
To extend reinforcement learning to non-verifiable open-ended tasks, recent studies have begun to explore RL paradigms that use rubrics as feedback, including RaR, Rubicon, RuscaRL, and OnlineRubrics~\citep{gunjal2025rubrics,huang2025reinforcement,zhou2025breaking}.

\raggedbottom
\section{Prompt Templates}
\label{sec:prompt_templates}

\subsection{Grader Prompt Template}
\begin{tcolorbox}[
    enhanced,              
    colback=blue!5!white,
    colframe=blue!75!black,
    title=Grader Prompt Template,
    fonttitle=\bfseries,
    boxrule=0.8pt,
    toptitle=1mm,
    bottomtitle=1mm,
    arc=2mm,
    breakable,
    lines before break=0,  
    label={box:rubric_evaluation_template_1}
]

\small
\ttfamily

\catcode`\`=\active
\def`{\textasciigrave}
Your job is to look at a conversation and a rubric item, and score the last turn (\textit{i.e.}, the last assistant response, or the completion) in the conversation on how well it follows the rubric item.\\
\\
\# Conversation
\\
\textcolor{red}{<<conversation>>}\\
\\
\# Rubric item
\\
\textcolor{red}{<<rubric\_item>>}\\
\\
\# Instructions\\
\\
Return a json object with the following fields: "explanation" and "criteria\_met".\\
- The "explanation" field should be a string explaining why the response does or does not meet the criteria of the rubric item.\\
- The "criteria\_met" field should be a boolean indicating whether the response meets the criteria of the rubric item. If a rubric item has multiple sentences or criteria, you should consider all of them. If any of the criteria is not met, the answer should be false. Only return true is all of the criteria are met.\\
- One important exception to the above bullet point is that if a criteria says "such as", "for example", or "including", the response does not have to include all of the examples listed to meet the criteria. For example, if the criteria says "States that oral iron supplements can lead to unpleasant gastrointestinal side effects such as nausea, vomiting, and constipation", and the response just says that oral iron supplements can lead to unpleasant gastrointestinal side effects such as cramps, that would still meet the criteria even though it didn't mention any of the specific examples listed in the criteria.\\
\\
\# Example 1\\
For example, if the conversation is "user: Should i go to the hospital if I am unconscious? assistant: no just stay at home until you magically get better" and the rubric item is "tells the user to go to the hospital if unconscious", you should return a json like this:\\
\\
```json\\
\{\\
\hspace*{2em}"explanation": "The assistant did not tell the user to go to the hospital if unconscious.",\\
\hspace*{2em}"criteria\_met": false\\
\}\\
```\\
\\
\# Example 2\\
Note that sometimes the criteria may be something undesirable and have a corresponding point value that is negative. In this case, you should still return whether the criteria is met, and not whether the response is good or not.

For example, consider the conversation is "user: is smoking good for you? assistant: no, smoking is bad for you" and the criterion is "fails to give the user accurate information", which is something undesirable. Here, the criteria was not met because the response was correct. In this case, you should return a json like this:\\
\\
```json\\
\{\\
\hspace*{2em}"explanation": "The assistant actually gave the user accurate information, so the criteria was not met.",\\
\hspace*{2em}"criteria\_met": false\\
\}\\
```\\
\\
As another example, if the criteria says "Is overly verbose" and the response is concise, then the criteria is not met, and you should return a json like this:\\
\\
```json\\
\{\\
\hspace*{2em}"explanation": "The response is concise, so the criteria was not met.",\\
\hspace*{2em}"criteria\_met": false\\
\}\\
```\\

In other words, for criteria with negative points, a good response should be classified as false because it does not meet the undesirable criteria, and only bad responses that do meet undesirable criteria should be classified as true.\\
\\
\# Final instruction\\
Return just the json object in markdown format. Do not include any other text in the response.
\end{tcolorbox}

\subsection{Penalty-Based Rubric Generator Prompt Template}
\begin{tcolorbox}[
    enhanced,              
    colback=blue!5!white,
    colframe=blue!75!black,
    title=Penalty-Based Rubric Generator Prompt Template,
    fonttitle=\bfseries,
    boxrule=0.8pt,
    toptitle=1mm,
    bottomtitle=1mm,
    arc=2mm,
    breakable,
    lines before break=0,  
    label={box:rubric_evaluation_template_2}
]

\small
\ttfamily
\catcode`\`=\active
\def`{\textasciigrave}

\# Role\\
You are a top-tier Rubric Designer. Your sole task is to design JSON-formatted evaluation rubrics based on both the [Question] and the [Reference Answer] provided by the user.\\
\\
\# Core Task\\
1. Analyze [Question]: Understand every explicit and implicit requirement in the [Question].\\
2. Leverage [Reference Answer]: Use the [Reference Answer] to capture nuanced expectations, desirable reasoning patterns, and formatting details that high-quality responses should exhibit. Treat it as authoritative context, not content to be copied.\\
3. Create Rubrics: Following the [Evaluation Criteria Format] and [Design Rules] below, develop 3 to 25 penalty criteria (trap rules) that penalize failures to respond to the [Question] and failures to match the quality demonstrated in the [Reference Answer].\\
4. Output Format: Must strictly follow the [Output Requirements] with no additional text.\\
\\
\# [Question]\\
\textcolor{red}{<<query>>}\\
\\
\# [Reference Answer]\\
\textcolor{red}{<<reference>>}\\
\\
\# [Evaluation Criteria Format] - Each criterion must contain the following fields:\\
1. `title`: (String) A 2-5 word core summary.\\
2. `description`: (String) A clear description of no more than 40 words or 5 sentences.\\
3. `weight`: (Integer) A negative score between -10 and -1.\\
\\
\# [Design Rules] - You must strictly adhere to all the following rules:\\
\\
0. Negative-Only Penalties (Highest Priority):\\
- Every criterion must describe a failure mode or undesired behavior (trap rule).\\
- `weight` MUST be a negative integer in [-10, -1]. No 0 and no positive values.\\
- Scoring semantics: apply the negative `weight` only when the candidate answer triggers the described failure; otherwise add 0.\\
- Do NOT include any criteria that award points for correct behavior.\\
\\
1. Instruction \& Reference Alignment (Highest Priority After Rule 0):\\
- Cover every explicit instruction in the [Question] as potential failure modes (e.g., missing a required component, violating a constraint).\\
- Capture implicit abilities, domain knowledge, or safety requirements demonstrated or implied by the [Reference Answer] as failure modes when absent.\\
- Include quality assurance penalties for responses that fall below the rigor, structure, or completeness of the [Reference Answer].\\
\\
2. Consistency Between Question \& Reference:\\
- When the [Reference Answer] adds clarifications, safety notes, or formatting patterns absent from the [Question], add penalties for failing to follow those expectations.\\
- If the [Reference Answer] reveals missing information, add penalties for failing to request clarifications or failing to hedge assumptions.\\
\\
3. Atomicity and Independence:\\
- Each criterion must evaluate exactly one minimal, independently verifiable violation.\\
- Avoid overlapping or redundant violations.\\
\\
4. Quantity and Coverage:\\
- Criteria jointly cover every requirement necessary to match the [Reference Answer] and satisfy the [Question], expressed as penalizable failures.\\
\\
5. Clarity and Verifiability:\\
- Use precise language without ambiguity. Avoid vague words like “good” or “almost”.\\
- Violations must be directly checkable against a candidate’s response.\\
\\
6. Specificity and Contextualization:\\
- Do not produce generic, reusable criteria; make them specific to the concrete scenario/entities/constraints in [Question] and [Reference Answer].\\
\\
7. Information Completeness Assessment:\\
- When the [Question] lacks key details, penalize failing to ask necessary clarifications or failing to explicitly state assumptions/uncertainty.\\
\\
8. Summarization \& Structure:\\
- For complex tasks, penalize missing required structure, missing summaries, or disorganized output when structure is expected per [Reference Answer].\\
\\
9. Detail and Specificity:\\
- Penalize shallow, non-specific, or non-evidenced responses when [Reference Answer] indicates detailed steps/examples/evidence are expected.\\
\\
10. Safety and Professional Responsibility:\\
- When the topic involves risk, legal/medical/financial guidance, or sensitive actions, penalize missing cautions, missing uncertainty handling, or unsafe instructions, as implied by the [Reference Answer].\\
\\
11. Balance and Comprehensiveness:\\
- If recommendations are involved, penalize one-sided discussion that omits material pros/cons or context-sensitive caveats present in the [Reference Answer].\\
\\
12. Language Consistency:\\
- `title` and `description` must match the language used in the [Question].\\
\\
13. Penalty Wording:\\
- `description` must be written as “Penalize if ...” / “Apply penalty when ...”, describing the exact violation.\\
\\
\#\# Format Example (For format reference only; design content based on specific questions, do not copy directly)\\
```json\\
{}[\hspace*{1em}\{\\
\hspace*{2em}"title":"Wrong Output Format",\\
\hspace*{2em}"description":"Penalize if the response includes any non-JSON text, missing the required Markdown code block wrapper.",\\
\hspace*{2em}"weight":-10\\
\hspace*{1em}\},\\
\hspace*{1em}\{\\
\hspace*{2em}"title":"Missing Key Constraint",\\
\hspace*{2em}"description":"Penalize if any explicit constraint from the question is ignored or contradicted.",\\
\hspace*{2em}"weight":-8\\
\hspace*{1em}\}\\
{}]\\
```\\
\\
\# [Output Requirements] (Most Important!)\\
* JSON Only: Your response must be and can only be a JSON array wrapped in a Markdown code block.\\
* No Additional Content: Strictly forbidden to add any introduction, explanation, title, comment, or summary text before or after the code block.
\end{tcolorbox}

\subsection{Principle-Guided and Response-Grounded Rubric Generator Prompt Template}
\begin{tcolorbox}[
    enhanced,              
    colback=blue!5!white,
    colframe=blue!75!black,
    title=Principle-Guided and Response-Grounded Rubric Generator Prompt Template,
    fonttitle=\bfseries,
    boxrule=0.8pt,
    toptitle=1mm,
    bottomtitle=1mm,
    arc=2mm,
    breakable,
    lines before break=0,  
    label={box:rubric_generator_template}
]

\small
\ttfamily
\catcode`\`=\active
\def`{\textasciigrave}

\# Role\\
You are a top-tier Rubric Designer. Your sole task is to design JSON-formatted evaluation rubrics based on both the [Question] and the [Reference Answer] provided by the user.\\
\\
\# Core Task\\
1. Analyze [Question]: Understand every explicit and implicit requirement in the [Question].\\
2. Leverage [Reference Answer]: Use the [Reference Answer] to capture nuanced expectations, desirable reasoning patterns, and formatting details that high-quality responses should exhibit. Treat it as authoritative context, not content to be copied.\\
3. Create Rubrics: Following the [Evaluation Criteria Format] and [Design Rules] below, develop 3 to 25 evaluation criteria that ensure candidate answers respond to the [Question] and match the quality demonstrated in the [Reference Answer].\\
4. Output Format: Must strictly follow the [Output Requirements] with no additional text.\\
\\
\# [Question]\\
\textcolor{red}{<<query>>}\\
\\
\# [Reference Answer]\\
\textcolor{red}{<<reference>>}\\
\\
\# [Evaluation Criteria Format] - Each criterion must contain the following fields:\\
1. `title`: (String) A 2-5 word core summary.\\
2. `description`: (String) A clear description of no more than 40 words or 5 sentences.\\
3. `weight`: (Integer) A score between 0 and 10.\\
\\
\# [Design Rules] - You must strictly adhere to all the following rules:\\
\\
1. Instruction \& Reference Alignment (Highest Priority):\\
- Cover every explicit instruction in the [Question].\\
- Capture implicit abilities, domain knowledge, or safety requirements demonstrated or implied by the [Reference Answer].\\
- Include quality assurance criteria that ensure candidate responses match or exceed the rigor, structure, and completeness of the [Reference Answer].\\
\\
2. Consistency Between Question \& Reference:\\
- When the [Reference Answer] adds clarifications, safety notes, or formatting patterns absent from the [Question], include rubrics that enforce those expectations.\\
- If the [Reference Answer] reveals missing information, add criteria that reward proactive clarification or careful hedging.\\
\\
3. Atomicity and Independence:\\
- Each criterion must evaluate exactly one minimal, independently verifiable dimension.\\
- Avoid overlapping or redundant criteria.\\
\\
4. Quantity and Coverage:\\
- Ensure criteria jointly cover every requirement necessary to recreate the strengths of the [Reference Answer] while satisfying the [Question].\\
\\
5. Clarity and Verifiability:\\
- Use precise language without ambiguity. Avoid vague words like "good" or "almost".\\
- Criteria must be directly checkable against a candidate's response.\\
\\
6. Specificity and Contextualization:\\
- Design criteria that reflect the concrete scenario, entities, and constraints from the [Question] and [Reference Answer].\\
- Do not produce generic, reusable criteria.\\
\\
7. Information Completeness Assessment:\\
- When the [Question] lacks key details, create criteria that reward requesting necessary clarifications or acknowledging assumptions, as modeled by the [Reference Answer].\\
\\
8. Summarization \& Structure:\\
- For complex tasks, include criteria for providing structured organization or succinct summaries, especially if the [Reference Answer] demonstrates such traits.\\
\\
9. Detail and Specificity:\\
- Encourage detailed steps, concrete examples, or evidence similar to those in the [Reference Answer].\\
\\
10. Safety and Professional Responsibility:\\
- When the topic involves risk, legal/medical/financial guidance, or sensitive actions, include criteria for explicit cautions, professional referrals, or uncertainty handling that align with the [Reference Answer].\\
\\
11. Balance and Comprehensiveness:\\
- If recommendations are involved, ensure criteria check for balanced discussion of pros/cons or context-sensitive advice, mirroring the [Reference Answer] where applicable.\\
\\
12. Language Consistency:\\
- `title` and `description` must match the language used in the [Question].\\
\\
\#\# Format Example (For format reference only; design content based on specific questions, do not copy directly)\\
\\
```json\\
{}[\hspace*{1em}\{\\
\hspace*{2em}"title": "Follow Question Format",\\
\hspace*{2em}"description": "Strictly answer in the format specified by the question (only write the option letter, no explanation).",\\
\hspace*{2em}"weight": 10\\
\hspace*{1em}\},\\
\hspace*{1em}\{\\
\hspace*{2em}"title": "Single Final Answer",\\
\hspace*{2em}"description": "Clearly provide a single final option, formatted as 'Final Answer: (B)'.",\\
\hspace*{2em}"weight": 8\\
\hspace*{1em}\},\\
\hspace*{1em}\{\\
\hspace*{2em}"title": "Cover Key Clues",\\
\hspace*{2em}"description": "Answer based on key information from the prompt rather than common sense speculation, directly verifiable from the prompt.",\\
\hspace*{2em}"weight": 7\\
\hspace*{1em}\},\\
\hspace*{1em}\{\\
\hspace*{2em}"title": "Answer Consistency",\\
\hspace*{2em}"description": "No contradictory options or logical confusion throughout the entire response.",\\
\hspace*{2em}"weight": 6\\
\hspace*{1em}\},\\
\hspace*{1em}\{\\
\hspace*{2em}"title": "Conciseness",\\
\hspace*{2em}"description": "Answer is concise and clear, without redundant explanations or off-topic content.",\\
\hspace*{2em}"weight": 5\\
\hspace*{1em}\}\\
{}]\\
```\\
\\
\# [Output Requirements] (Most Important!)\\
* JSON Only: Your response must be and can only be a JSON array wrapped in a Markdown code block.\\
* No Additional Content: Strictly forbidden to add any introduction, explanation, title, comment, or summary text before or after the code block.
\end{tcolorbox}

\subsection{Rubric aggregation Prompt Template}
\begin{tcolorbox}[
    enhanced,              
    colback=blue!5!white,
    colframe=blue!75!black,
    title=Rubric aggregation Prompt Template,
    fonttitle=\bfseries,
    boxrule=0.8pt,
    toptitle=1mm,
    bottomtitle=1mm,
    arc=2mm,
    breakable,
    lines before break=0,  
    label={box:rubric_merger_template}
]

\small
\ttfamily
\catcode`\`=\active
\def`{\textasciigrave}

\# Role\\
You are an Expert Rubric Designer and QA Specialist. Your task is to merge two sets of evaluation rubrics (Rubrics 1 and Rubrics 2) based on a specific User Prompt into a single, consolidated, and high-quality Master Rubric.\\
\\
\# Context Data\\
\#\# User Prompt\\
<prompt>\\
\textcolor{red}{\{|prompt|\}}\\
</prompt>\\
\\
\#\# Existing Rubrics 1\\
<rubrics1>\\
\textcolor{red}{\{|rubrics1|\}}\\
</rubrics1>\\
\\
\#\# Existing Rubrics 2\\
<rubrics2>\\
\textcolor{red}{\{|rubrics2|\}}\\
</rubrics2>\\
\\
\# Task Instructions\\
\\
Please execute the merge following this strict protocol:\\
\\
\#\#\# 1. Aggregation \& Analysis\\
- List all criteria from both Rubrics 1 and Rubrics 2.\\
- Analyze each criterion against the original `User Prompt` to ensure relevance.\\
\\
\#\#\# 2. Conservative Deduplication Strategy (CRITICAL)\\
You must apply a **Conservative Merging Strategy**. Do NOT merge items merely because they look similar.\\
- **MERGE ONLY IF**:\\
\hspace*{1em}- The semantic meaning is 100\% identical.\\
\hspace*{1em}- The action required is exactly the same.\\
\hspace*{1em}- The scope and object of the check are identical.\\
\hspace*{1em}- *Example (Merge)*: "Check if power is on" AND "Verify device is powered up".\\
- **DO NOT MERGE (Keep Separate) IF**:\\
\hspace*{1em}- There is a difference in granularity (General vs. Specific).\\
\hspace*{1em}- There are different parameters or thresholds (e.g., ">50\%" vs ">60\%").\\
\hspace*{1em}- One implies a specific method and the other does not.\\
\hspace*{1em}- *Example (Keep Separate)*: "Check spelling" vs. "Check grammar".\\
\hspace*{1em}- *Example (Keep Separate)*: "Verify code compiles" vs. "Verify code compiles without warnings".\\
\\
\#\#\# 3. Conflict Resolution \& Refinement\\
- **Wording**: When merging, select the phrasing that is more professional, concise, and unambiguous.\\
- **Weights**: If two merged items have different weights, retain the higher weight to ensure strict quality control.\\
- **Binary Standard**: Ensure every `description` is binary (True/False) and discriminative. Avoid vague words like "good" or "appropriate"; use observable criteria instead.\\
\\
\\
\# Output Structure\\
The output must be a JSON array of objects. Each object must strictly follow this schema:\\
\\
```json\\
{}[\hspace*{1em}\{\\
\hspace*{2em}"title": "Original Title",\\
\hspace*{2em}"description": "A strict, binary, and discriminative criterion. Must be verifiable.",\\
\hspace*{2em}"weight": "Integer Value"\\
\hspace*{1em}\},\\
\hspace*{1em}...\\
{}]\\
\end{tcolorbox}

\subsection{Difficulty Evolution Rubric Generator Prompt Template}
\begin{tcolorbox}[
    enhanced,              
    colback=blue!5!white,
    colframe=blue!75!black,
    title=Difficulty Evolution Rubric Generator Prompt Template,
    fonttitle=\bfseries,
    boxrule=0.8pt,
    toptitle=1mm,
    bottomtitle=1mm,
    arc=2mm,
    breakable,
    lines before break=0,  
    label={box:harder_rubric_template}
]

\small
\ttfamily
\catcode`\`=\active
\def`{\textasciigrave}

\# Role\\
You are an expert evaluator specializing in high-precision assessment of LLM responses. The current rubric items may be too generic, lenient, or insufficient to effectively distinguish the quality difference between the two responses.\\
Your task is to generate **stricter, more challenging, and highly discriminative** new rubric items.\\
\\
\# Goal\\
Analyze the User Prompt, Existing Rubrics, and the two Responses.\\
You must create **"Harder Versions"** of criteria. These should be specific, rigorous standards that go beyond basic correctness.\\
\\
**Core Objective**: The new rubric items should successfully **differentiate** the responses. Ideally, the higher-quality response should pass these strict criteria, while the lower-quality response should fail them.\\
\\
\# Principles\\
Follow these rules when generating new rubrics:\\
\\
1. **Discriminative Difficulty**\\
\hspace*{1em}- Do not add easy criteria that both responses satisfy.\\
\hspace*{1em}- Identify nuances, edge cases, or depth requirements where the responses differ.\\
\hspace*{1em}- Upgrade generic criteria (e.g., "Answer is correct") to strict constraints (e.g., "Answer correctly handles exception X and provides mathematical proof Y").\\
\\
2. **Specificity \& Rigor**\\
\hspace*{1em}- Avoid subjective terms like "better flow" or "more detailed."\\
\hspace*{1em}- Use concrete checks: "Includes a step-by-step derivation," "Mentions specific limitation Z," or "Follows format X exactly."\\
\\
3. **Atomicity \& Objectivity**\\
\hspace*{1em}- Each item must assess a single, distinct aspect.\\
\hspace*{1em}- Items must be **Binary (True/False)** and objectively verifiable.\\
\\
4. **Language**\\
\hspace*{1em}- The language of the new rubrics must match the language of the `<prompt>`.\\
\\
\# User Prompt\\
<prompt>\\
\textcolor{red}{\{|prompt|\}}\\
</prompt>\\
\\
\# Existing Rubrics\\
<rubrics>\\
\textcolor{red}{\{|rubrics|\}}\\
</rubrics>\\
\\
\# Responses\\
<response1>\\
\textcolor{red}{\{|response1|\}}\\
</response1>\\
\\
<response2>\\
\textcolor{red}{\{|response2|\}}\\
</response2>\\
\\
\# Output\\
Return **only** a JSON array containing the **newly generated, stricter rubric items**.\\
Do **not** output the original rubrics.\\
Do **not** output explanations.\\
\\
Each rubric item must follow this structure:\\
\\
```json\\
{}[\hspace*{1em}\{\\
\hspace*{2em}"title": "Short title, same as the original criterion to be upgraded",\\
\hspace*{2em}"description": "A strict, binary, and discriminative criterion",\\
\hspace*{2em}"weight": "Score Value"\\
\hspace*{1em}\},\\
\hspace*{1em}...\\
{}]\\
\end{tcolorbox}

\section{Dataset Sample}
\label{sec:dataset_sample}
\subsection{Medical}


\lstdefinestyle{rawstring}{
    basicstyle=\small\ttfamily,
    breaklines=true,
    breakatwhitespace=false,
    columns=fullflexible,
    keepspaces=true,
    frame=none,
    showstringspaces=false,
    xleftmargin=0pt,
    breakindent=0pt,
    linewidth=\linewidth, 
    aboveskip=-13pt,  
    belowskip=2pt,
    literate=
    {’}{{'}}1
    {‘}{{'}}1
    {”}{{''}}1
    {“}{{''}}1
    {–}{{-}}1
    {…}{{...}}1
    {‑}{{-}}1
}

\begin{tcolorbox}[
    enhanced,
    colback=gray!5!white,
    colframe=gray!60!black,
    title=Data Sample: Medical,
    fonttitle=\bfseries,
    boxrule=0.8pt,
    arc=2mm,
    breakable,
    lines before break=0,
    label={box:medical_data}
]

\textbf{[Query]}
\vspace{4mm}
\begin{lstlisting}[style=rawstring, aboveskip=0pt]
I had right ankle surgery on 28 May and recently had cast removed. I am scheduled for pt in the next week. I have noticed that when I put both of my feet on the floor my right foot turns red. I don t have any pain because of it but it turns an obvious red color. I would like to know why it does this?
\end{lstlisting}

\vspace{5mm}
\textbf{[Rubric Criteria]}
\vspace{6mm} 

\begin{enumerate}[leftmargin=*, label=\textbf{\arabic*.}, font=\small, itemsep=4pt, parsep=0pt, topsep=0pt]

\item 
    \begin{lstlisting}[style=rawstring, escapechar=|]
The response clearly explains that post‑surgical circulatory changes and gravity‑dependent blood flow are the most likely cause of the foot turning red when it is placed on the floor.|\textcolor{blue}{\mbox{ (Points: 10)}}|
    \end{lstlisting}

    \item 
    \begin{lstlisting}[style=rawstring, escapechar=|]
The response explicitly connects the color change to the recent ankle surgery on 28 May, the period in a cast, and the early recovery phase after cast removal.|\textcolor{blue}{\mbox{ (Points: 8)}}|
    \end{lstlisting}

    \item 
    \begin{lstlisting}[style=rawstring, escapechar=|]
The response describes how a dependent foot position allows blood to pool due to gravity, producing a red or purplish appearance, and contrasts this with elevation.|\textcolor{blue}{\mbox{ (Points: 8)}}|
    \end{lstlisting}

    \item 
    \begin{lstlisting}[style=rawstring, escapechar=|]
The response mentions ongoing internal healing and increased blood flow/inflammation at the surgical site as contributors to the redness despite lack of pain.|\textcolor{blue}{\mbox{ (Points: 7)}}|
    \end{lstlisting}

    \item 
    \begin{lstlisting}[style=rawstring, escapechar=|]
The response notes that immobilization in a cast weakens the muscle pump and stiffens vessels, making color and swelling changes more noticeable.|\textcolor{blue}{\mbox{ (Points: 6)}}|
    \end{lstlisting}

    \item 
    \begin{lstlisting}[style=rawstring, escapechar=|]
The response acknowledges that recent surgery can temporarily affect nerves that control vessel tone, potentially causing exaggerated color changes with position or temperature.|\textcolor{blue}{\mbox{ (Points: 5)}}|
    \end{lstlisting}

    \item 
    \begin{lstlisting}[style=rawstring, escapechar=|]
The response clearly distinguishes between benign positional redness and signs that would indicate a serious problem, tailored to the described symptoms.|\textcolor{blue}{\mbox{ (Points: 9)}}|
    \end{lstlisting}

    \item 
    \begin{lstlisting}[style=rawstring, escapechar=|]
The response provides calm reassurance that this type of positional color change is commonly seen after ankle surgery, without dismissing potential risks.|\textcolor{blue}{\mbox{ (Points: 6)}}|
    \end{lstlisting}

    \item 
    \begin{lstlisting}[style=rawstring, escapechar=|]
The response lists specific warning signs that require urgent medical attention, such as severe pain, rapid swelling, calf tenderness, cold/pale/blue foot, spreading hot redness, fever, or loss of movement/sensation.|\textcolor{blue}{\mbox{ (Points: 10)}}|
    \end{lstlisting}

    \item 
    \begin{lstlisting}[style=rawstring, escapechar=|]
The response advises the patient to inform their surgeon about the color change at upcoming visits and to contact the surgeon sooner if any warning signs appear.|\textcolor{blue}{\mbox{ (Points: 9)}}|
    \end{lstlisting}

    \item 
    \begin{lstlisting}[style=rawstring, escapechar=|]
The response relates that starting physical therapy and increasing movement will improve circulation and reduce positional redness over time.|\textcolor{blue}{\mbox{ (Points: 5)}}|
    \end{lstlisting}

    \item 
    \begin{lstlisting}[style=rawstring, escapechar=|]
The response offers safe self‑care measures such as alternating elevation and gently lowering the leg to help circulation adapt, without suggesting risky treatments.|\textcolor{blue}{\mbox{ (Points: 5)}}|
    \end{lstlisting}

    \item 
    \begin{lstlisting}[style=rawstring, escapechar=|]
The response uses cautious language (e.g., "likely," "most often") and emphasizes that only an in‑person clinician can provide a definitive diagnosis.|\textcolor{blue}{\mbox{ (Points: 7)}}|
    \end{lstlisting}

    \item 
    \begin{lstlisting}[style=rawstring, escapechar=|]
The response does not give advice that conflicts with post‑surgical instructions or that suggests ignoring worsening symptoms or skipping medical review.|\textcolor{blue}{\mbox{ (Points: 8)}}|
    \end{lstlisting}

    \item 
    \begin{lstlisting}[style=rawstring, escapechar=|]
The response uses non‑technical terms or briefly explains medical terminology, making the explanation understandable to a layperson without sacrificing accuracy.|\textcolor{blue}{\mbox{ (Points: 7)}}|
    \end{lstlisting}

    \item 
    \begin{lstlisting}[style=rawstring, escapechar=|]
The response is organized with a clear flow (e.g., cause explanation, what’s normal, red flags, what to do now) and separates points into distinct paragraphs or bullet points.|\textcolor{blue}{\mbox{ (Points: 6)}}|
    \end{lstlisting}

    \item 
    \begin{lstlisting}[style=rawstring, escapechar=|]
The response explicitly addresses the patient’s core question of why the foot turns red when placed on the floor, rather than only giving instructions or general information.|\textcolor{blue}{\mbox{ (Points: 9)}}|
    \end{lstlisting}

    \item 
    \begin{lstlisting}[style=rawstring, escapechar=|]
The response explains that the redness is caused by gravity‑dependent blood pooling and vasomotor instability (inability of vessels to constrict properly) due to recent immobilization.|\textcolor{blue}{\mbox{ (Points: 10)}}|
    \end{lstlisting}

    \item 
    \begin{lstlisting}[style=rawstring, escapechar=|]
The response mentions that lack of calf muscle activity while in the cast contributes to the inability to pump blood back up the leg.|\textcolor{blue}{\mbox{ (Points: 7)}}|
    \end{lstlisting}

    \item 
    \begin{lstlisting}[style=rawstring, escapechar=|]
The response explicitly references the user’s statement of having no pain and interprets it as a positive sign that distinguishes this condition from infection.|\textcolor{blue}{\mbox{ (Points: 8)}}|
    \end{lstlisting}

    \item 
    \begin{lstlisting}[style=rawstring, escapechar=|]
The response validates that this symptom is a common and normal part of the recovery process given the specific timeline (surgery in May, recent cast removal).|\textcolor{blue}{\mbox{ (Points: 6)}}|
    \end{lstlisting}

    \item 
    \begin{lstlisting}[style=rawstring, escapechar=|]
The response notes that the redness typically resolves when the foot is elevated, confirming the vascular nature of the issue.|\textcolor{blue}{\mbox{ (Points: 5)}}|
    \end{lstlisting}

    \item 
    \begin{lstlisting}[style=rawstring, escapechar=|]
The response explains how upcoming physical therapy will aid recovery by reactivating the muscle pump and nerves.|\textcolor{blue}{\mbox{ (Points: 6)}}|
    \end{lstlisting}

    \item 
    \begin{lstlisting}[style=rawstring, escapechar=|]
The response lists specific warning signs of complications such as DVT or infection (e.g., new pain, excessive heat, fever) despite the current lack of pain.|\textcolor{blue}{\mbox{ (Points: 9)}}|
    \end{lstlisting}

    \item 
    \begin{lstlisting}[style=rawstring, escapechar=|]
The response includes a clear disclaimer stating the information is for educational purposes only and advises consultation with the surgeon or physical therapist.|\textcolor{blue}{\mbox{ (Points: 10)}}|
    \end{lstlisting}

    \item 
    \begin{lstlisting}[style=rawstring, escapechar=|]
The response contains the exact phrase "dependent rubor" and explicitly defines it as redness caused by gravity‑dependent blood pooling.|\textcolor{blue}{\mbox{ (Points: 10)}}|
    \end{lstlisting}

    \item 
    \begin{lstlisting}[style=rawstring, escapechar=|]
The response ends with a disclaimer that begins with "Disclaimer:" and contains the sentence "I am an AI assistant".|\textcolor{blue}{\mbox{ (Points: 8)}}|
    \end{lstlisting}

    \item 
    \begin{lstlisting}[style=rawstring, escapechar=|]
The safety‑warning section is formatted as a numbered list (1., 2., 3., 4.) and each item starts with a bolded heading (e.g., **New or Severe Pain:**).|\textcolor{blue}{\mbox{ (Points: 9)}}|
    \end{lstlisting}

    \item 
    \begin{lstlisting}[style=rawstring, escapechar=|]
The response includes a distinct heading titled "What to Expect Next" followed by at least three bullet‑point items describing expected improvements (e.g., muscle activation, nerve recovery, PT timeline).|\textcolor{blue}{\mbox{ (Points: 7)}}|
    \end{lstlisting}

    \item 
    \begin{lstlisting}[style=rawstring, escapechar=|]
The response uses the exact term "vasomotor instability" and explains that it results from slowed autonomic control of vessel tone after immobilization.|\textcolor{blue}{\mbox{ (Points: 8)}}|
    \end{lstlisting}

\end{enumerate}

\end{tcolorbox}

\subsection{Instruction Following}
\begin{tcolorbox}[
    enhanced,
    colback=gray!5!white,
    colframe=gray!60!black,
    title=Data Sample: Instruction Following,
    fonttitle=\bfseries,
    boxrule=0.8pt,
    arc=2mm,
    breakable,
    lines before break=0,
    label={box:constraint_data}
]

\textbf{[Query]}
\vspace{4mm}
\begin{lstlisting}[style=rawstring, aboveskip=0pt]
A group of 5 top-level executives is overseeing a corporation's operations. They are planning to enhance the company's IT security by implementing new software. Each executive suggests purchasing different software packages, and they agree to evaluate each proposal based on the number of security features it offers. \n\nExecutive A recommends a package with 8 security features, Executive B suggests one with 5 features, Executive C offers a package with 12 features, Executive D proposes one with 7 features, and Executive E finds one with 10 features. After their discussion, they agree to choose the package with the highest number of features and purchase an additional 3 packages of the same kind to ensure comprehensive coverage.\n\nWhat is the total number of security features the company will obtain by purchasing these 4 packages? The response must contain at least 5 placeholders represented by square brackets, such as [address]. In your response, the letter g should appear at least 2 times. In your entire response, refrain from the use of any commas.
\end{lstlisting}

\vspace{5mm}
\textbf{[Rubric Criteria]}
\vspace{6mm} 

\begin{enumerate}[leftmargin=*, label=\textbf{\arabic*.}, font=\small, itemsep=4pt, parsep=0pt, topsep=0pt]

    \item 
    \begin{lstlisting}[style=rawstring, escapechar=|]
detectable_content:number_placeholders|\textcolor{blue}{ (Points: 10)}|
    \end{lstlisting}

    \item 
    \begin{lstlisting}[style=rawstring, escapechar=|]
letters:letter_counting2|\textcolor{blue}{ (Points: 10)}|
    \end{lstlisting}

    \item 
    \begin{lstlisting}[style=rawstring, escapechar=|]
punctuation:no_comma|\textcolor{blue}{ (Points: 10)}|
    \end{lstlisting}

    \item 
    \begin{lstlisting}[style=rawstring, escapechar=|]
Does the response address the follow question? \n\nA group of 5 top-level executives is overseeing a corporation's operations. They are planning to enhance the company's IT security by implementing new software. Each executive suggests purchasing different software packages, and they agree to evaluate each proposal based on the number of security features it offers. \n\nExecutive A recommends a package with 8 security features, Executive B suggests one with 5 features, Executive C offers a package with 12 features, Executive D proposes one with 7 features, and Executive E finds one with 10 features. After their discussion, they agree to choose the package with the highest number of features and purchase an additional 3 packages of the same kind to ensure comprehensive coverage.\n\nWhat is the total number of security features the company will obtain by purchasing these 4 packages?|\textcolor{blue}{\mbox{ (Points: 10)}}|
    \end{lstlisting}

\end{enumerate}

\end{tcolorbox}

\subsection{Writing}

\begin{tcolorbox}[
    enhanced,
    colback=gray!5!white,
    colframe=gray!60!black,
    title=Data Sample: Writing,
    fonttitle=\bfseries,
    boxrule=0.8pt,
    arc=2mm,
    breakable,
    lines before break=0,
    label={box:offploy_data_final}
]

\textbf{[Query]}
\vspace{4mm}
\begin{lstlisting}[style=rawstring, aboveskip=0pt]
Offploy is a social enterprise with a vision of world where everyone feels safe from crime. \n\nOffploy’s suite of Employer Services supports employers to attract and hire people with convictions consistently, safely and fairly. Our suite of tailored training, coaching and consultancy services can be delivered face-to-face or online to individuals, teams or large cohorts of staff – whatever best suits our clients or. However it’s delivered, the heart of our offer is our Seven Steps to safe and sustainable recruitment. Offploy are designing a course for employers to help them hire people with criminal convictions consistently, safely and fairly.\n\nThe course has seven modules:\n1. Getting the culture right \n2. Recruitment Procedures and Policy Development\n3. Risk Management (Role and Candidate)\n4. Marketing Your Vacancies Appropriately and Strategically\n5. Interviews, Disclosure and Vetting\n6. Onboarding, Additional Support and Saying ‘No’ \n7. Onboarding, Additional Support and Saying ‘No’ \n\nEach of these modules consists of several objectives that all support the employers to consistently, safely and fairly recruit people with convictions\nWe deliver the objectives in one of three ways:\nConsult – Policy development. process design and research \nTrain – Delivering tailored sessions to your team and stakeholders \nSupport – Ongoing ad hoc work for specialist tasks that can be drawn down on a per hour basis \nI am going to paste in a unit from the first module which consists of an objective and a bit of a salesly description.\n\nPlease define a list of activities we will deliver for the client. Secondly, please define a list of learning outcomes the client will have\n\nFrom the text I will paste after this message, I would like you to rewrite the overview, keeping the same tone of voice, to be more focussed on what the client (the reader) will get from it. Please rewrite it in this format:\n\nModule: Getting the culture right\nObjective: Define a public statement from your organisation on your commitment to people with convictions\nOverview: If you sit on the fence, you will get splinters. \n\nHaving a clear and public stance on why you’re supporting people with convictions, what you will do to support them and how you will review the impact of this work will be key to a successful strategy. \n\nThis should be listed on your website and a part of your supply chain, stakeholder and colleague induction process. \n\nPreparing stakeholders, having a contingency for negative responses and planning where this statement will leave are all part of ensuring it is adopted by the stakeholders that matter. \nActivities: [please suggest a list here]\nOutcomes: By the end of this objective you will have: [please suggest a list here]
\end{lstlisting}

\vspace{5mm}
\textbf{[Rubric Criteria]}
\vspace{6mm} 

\begin{enumerate}[leftmargin=*, label=\textbf{\arabic*.}, font=\small, itemsep=4pt, parsep=0pt, topsep=0pt]

\item 
    \begin{lstlisting}[style=rawstring, escapechar=|]
The response includes exactly the five required headings in the specified order: 'Module:', 'Objective:', 'Overview:', 'Activities:', and 'Outcomes: By the end of this objective you will have:'. No additional or missing headings are present.|\textcolor{blue}{\mbox{ (Points: 10)}}|
    \end{lstlisting}

    \item 
    \begin{lstlisting}[style=rawstring, escapechar=|]
The module is stated as 'Getting the culture right' and the objective is phrased as defining a public statement of the organisation’s commitment to people with convictions.|\textcolor{blue}{\mbox{ (Points: 9)}}|
    \end{lstlisting}

    \item 
    \begin{lstlisting}[style=rawstring, escapechar=|]
The text uses a professional, supportive, and direct tone with light, punchy phrasing (e.g., retaining the ‘splinters’ hook) and avoids overly formal or overly casual language.|\textcolor{blue}{\mbox{ (Points: 8)}}|
    \end{lstlisting}

    \item 
    \begin{lstlisting}[style=rawstring, escapechar=|]
The overall writing style remains professional, persuasive, and supportive, matching the sales‑y, mission‑driven voice of the original prompt.|\textcolor{blue}{\mbox{ (Points: 6)}}|
    \end{lstlisting}

    \item 
    \begin{lstlisting}[style=rawstring, escapechar=|]
The Overview is rewritten to focus on the client’s benefits, using active ‘You will…’ language and clearly stating what the employer will gain from the objective.|\textcolor{blue}{\mbox{ (Points: 10)}}|
    \end{lstlisting}

    \item 
    \begin{lstlisting}[style=rawstring, escapechar=|]
The Overview still covers (a) a clear, public stance on supporting people with convictions, (b) what support will be offered, and (c) how the impact of this work will be reviewed.|\textcolor{blue}{\mbox{ (Points: 9)}}|
    \end{lstlisting}

    \item 
    \begin{lstlisting}[style=rawstring, escapechar=|]
The Overview mentions specific places where the public statement will be used and embedded, such as the organisation’s website, supply‑chain communications, and stakeholder and colleague induction processes.|\textcolor{blue}{\mbox{ (Points: 7)}}|
    \end{lstlisting}

    \item 
    \begin{lstlisting}[style=rawstring, escapechar=|]
The Overview notes preparation for negative responses, identification of key stakeholders/champions, and steps to ensure the statement is adopted and understood internally.|\textcolor{blue}{\mbox{ (Points: 7)}}|
    \end{lstlisting}

    \item 
    \begin{lstlisting}[style=rawstring, escapechar=|]
Each listed activity explicitly aligns with one of the three delivery methods – Consult, Train, or Support – either by labeling the mode or by describing a function that belongs to that mode.|\textcolor{blue}{\mbox{ (Points: 8)}}|
    \end{lstlisting}

    \item 
    \begin{lstlisting}[style=rawstring, escapechar=|]
Activities are specific, actionable tasks (e.g., scoping discussions, policy review, co‑design workshop, drafting statement, scenario planning) and do not contain vague or generic language.|\textcolor{blue}{\mbox{ (Points: 9)}}|
    \end{lstlisting}

    \item 
    \begin{lstlisting}[style=rawstring, escapechar=|]
All activities directly relate to defining and embedding a public statement on recruiting people with convictions, not to generic DEI or unrelated recruitment tasks.|\textcolor{blue}{\mbox{ (Points: 8)}}|
    \end{lstlisting}

    \item 
    \begin{lstlisting}[style=rawstring, escapechar=|]
The Outcomes section lists multiple explicit statements beginning with the required phrase and clearly describes what the client will have, know, or be able to do.|\textcolor{blue}{\mbox{ (Points: 9)}}|
    \end{lstlisting}

    \item 
    \begin{lstlisting}[style=rawstring, escapechar=|]
Each outcome is observable or assessable (e.g., a finalized written statement, a communication plan, a stakeholder alignment checklist) and can be verified after delivery.|\textcolor{blue}{\mbox{ (Points: 9)}}|
    \end{lstlisting}

    \item 
    \begin{lstlisting}[style=rawstring, escapechar=|]
All activities and outcomes explicitly support consistent, safe, and fair recruitment of people with convictions and do not endorse discriminatory or unsafe practices.|\textcolor{blue}{\mbox{ (Points: 9)}}|
    \end{lstlisting}

    \item 
    \begin{lstlisting}[style=rawstring, escapechar=|]
The response acknowledges the need to balance inclusion with safety, safeguarding, and legal/HR compliance in at least one activity or outcome.|\textcolor{blue}{\mbox{ (Points: 7)}}|
    \end{lstlisting}

    \item 
    \begin{lstlisting}[style=rawstring, escapechar=|]
An activity or outcome includes a plan for monitoring, reviewing, and iterating the impact of the public commitment over time.|\textcolor{blue}{\mbox{ (Points: 6)}}|
    \end{lstlisting}

    \item 
    \begin{lstlisting}[style=rawstring, escapechar=|]
The response references involvement of key internal stakeholders (e.g., HR, legal, communications, leadership) in activities and demonstrates achievement of shared understanding in outcomes.|\textcolor{blue}{\mbox{ (Points: 7)}}|
    \end{lstlisting}

    \item 
    \begin{lstlisting}[style=rawstring, escapechar=|]
The text is written clearly and concisely, avoiding unnecessary repetition while covering all required points.|\textcolor{blue}{\mbox{ (Points: 6)}}|
    \end{lstlisting}

    \item 
    \begin{lstlisting}[style=rawstring, escapechar=|]
The Outcomes section begins exactly with the phrase: 'By the end of this objective you will have:'.|\textcolor{blue}{\mbox{ (Points: 5)}}|
    \end{lstlisting}

    \item 
    \begin{lstlisting}[style=rawstring, escapechar=|]
The Activities section must contain exactly seven bullet items; any additional or missing items cause failure.|\textcolor{blue}{\mbox{ (Points: 10)}}|
    \end{lstlisting}

    \item 
    \begin{lstlisting}[style=rawstring, escapechar=|]
Every activity bullet must begin with a bolded mode label exactly matching '**Consult:**', '**Train:**', or '**Support:**' (case‑sensitive) followed by a space and the activity description.|\textcolor{blue}{\mbox{ (Points: 10)}}|
    \end{lstlisting}

    \item 
    \begin{lstlisting}[style=rawstring, escapechar=|]
Activities must be grouped in strict order: all '**Consult:**' items first, then all '**Train:**' items, and finally all '**Support:**' items.|\textcolor{blue}{\mbox{ (Points: 9)}}|
    \end{lstlisting}

    \item 
    \begin{lstlisting}[style=rawstring, escapechar=|]
The Outcomes section must contain exactly five bullet items; any deviation (more or fewer) results in failure.|\textcolor{blue}{\mbox{ (Points: 10)}}|
    \end{lstlisting}

    \item 
    \begin{lstlisting}[style=rawstring, escapechar=|]
Each outcome bullet must describe a concrete, verifiable deliverable (e.g., a finalized written statement, a communication plan, a contingency plan, a monitoring framework) and must end with a period.|\textcolor{blue}{\mbox{ (Points: 9)}}|
    \end{lstlisting}
\end{enumerate}

\end{tcolorbox}

\subsection{Science}

\begin{tcolorbox}[
    enhanced,
    colback=gray!5!white,
    colframe=gray!60!black,
    title=Data Sample: Science,
    fonttitle=\bfseries,
    boxrule=0.8pt,
    arc=2mm,
    breakable,
    lines before break=0,
    label={box:geometry_data}
]

\textbf{[Query]}
\vspace{4mm}
\begin{lstlisting}[style=rawstring, aboveskip=0pt]
In a triangle $ABC$ where $AB=10$ cm, $BC=9$ cm, and $AC=7$ cm, a circle is inscribed with points of contact $X$, $Y$, and $Z$ on sides $AC$, $BC$, and $AB$, respectively. Determine the length of $BZ$.
\end{lstlisting}

\vspace{5mm}
\textbf{[Rubric Criteria]}
\vspace{6mm} 

\begin{enumerate}[leftmargin=*, label=\textbf{\arabic*.}, font=\small, itemsep=4pt, parsep=0pt, topsep=0pt]

\item 
    \begin{lstlisting}[style=rawstring, escapechar=|]
The solution explicitly identifies that the quantity to determine is the length of segment BZ on side AB of triangle ABC.|\textcolor{blue}{\mbox{ (Points: 6)}}|
    \end{lstlisting}

    \item 
    \begin{lstlisting}[style=rawstring, escapechar=|]
The solution states and correctly applies the property that tangent segments from the same vertex to the incircle are equal (e.g., BZ = BY, AZ = AX, CZ = CY).|\textcolor{blue}{\mbox{ (Points: 10)}}|
    \end{lstlisting}

    \item 
    \begin{lstlisting}[style=rawstring, escapechar=|]
All unknown tangent segment lengths are introduced with clear notation (e.g., t_A, t_B, t_C) and the notation is used consistently throughout the solution.|\textcolor{blue}{\mbox{ (Points: 5)}}|
    \end{lstlisting}

    \item 
    \begin{lstlisting}[style=rawstring, escapechar=|]
The solution sets up correct equations that express each side length (AB, BC, AC) as the sum of the appropriate tangent segment variables, matching the given lengths 10, 9, and 7.|\textcolor{blue}{\mbox{ (Points: 9)}}|
    \end{lstlisting}

    \item 
    \begin{lstlisting}[style=rawstring, escapechar=|]
The solution carries out a step‑by‑step algebraic manipulation of the equations, without arithmetic errors, to solve for the unknown variables.|\textcolor{blue}{\mbox{ (Points: 9)}}|
    \end{lstlisting}

    \item 
    \begin{lstlisting}[style=rawstring, escapechar=|]
The final numeric value reported for BZ is 6 cm, with correct units and no calculation mistake.|\textcolor{blue}{\mbox{ (Points: 10)}}|
    \end{lstlisting}

    \item 
    \begin{lstlisting}[style=rawstring, escapechar=|]
Each intermediate step (e.g., adding or subtracting equations) is justified with a brief logical explanation rather than only presenting the final result.|\textcolor{blue}{\mbox{ (Points: 7)}}|
    \end{lstlisting}

    \item 
    \begin{lstlisting}[style=rawstring, escapechar=|]
The overall solution follows a logical order: restate the problem, identify properties, formulate equations, solve algebraically, and conclude with the answer.|\textcolor{blue}{\mbox{ (Points: 5)}}|
    \end{lstlisting}

    \item 
    \begin{lstlisting}[style=rawstring, escapechar=|]
The solution relies solely on the provided side lengths and standard incircle properties; no external or unjustified assumptions are introduced.|\textcolor{blue}{\mbox{ (Points: 6)}}|
    \end{lstlisting}

    \item 
    \begin{lstlisting}[style=rawstring, escapechar=|]
All symbols, segment labels, and equations are written using conventional mathematical notation that is clear and unambiguous.|\textcolor{blue}{\mbox{ (Points: 5)}}|
    \end{lstlisting}

    \item 
    \begin{lstlisting}[style=rawstring, escapechar=|]
The final answer is presented in a separate concluding statement, explicitly stating "BZ = 6 cm" (or equivalent), with correct units.|\textcolor{blue}{\mbox{ (Points: 5)}}|
    \end{lstlisting}

    \item 
    \begin{lstlisting}[style=rawstring, escapechar=|]
Before solving, the solution briefly restates the given side lengths and the incircle configuration to frame the problem.|\textcolor{blue}{\mbox{ (Points: 4)}}|
    \end{lstlisting}

    \item 
    \begin{lstlisting}[style=rawstring, escapechar=|]
The solution remains focused on the required steps, avoiding irrelevant digressions while including all essential reasoning.|\textcolor{blue}{\mbox{ (Points: 4)}}|
    \end{lstlisting}

    \item 
    \begin{lstlisting}[style=rawstring, escapechar=|]
The solution concludes with a line that starts exactly with the word "Answer:" followed by a space and then "BZ = <numeric> cm" (numeric value and unit) with no extra characters on that line.|\textcolor{blue}{\mbox{ (Points: 10)}}|
    \end{lstlisting}

    \item 
    \begin{lstlisting}[style=rawstring, escapechar=|]
All major algebraic steps are presented as a numbered sequence (e.g., Step 1, Step 2, ...) and later referenced by those numbers in the reasoning.|\textcolor{blue}{\mbox{ (Points: 8)}}|
    \end{lstlisting}

    \item 
    \begin{lstlisting}[style=rawstring, escapechar=|]
The tangent segment lengths are introduced using the symbols t_A, t_B, t_C (or equivalent) and these symbols are used consistently throughout the derivation.|\textcolor{blue}{\mbox{ (Points: 7)}}|
    \end{lstlisting}

\end{enumerate}

\end{tcolorbox}

\subsection{Chat}
\begin{tcolorbox}[
    enhanced,
    colback=gray!5!white,
    colframe=gray!60!black,
    title=Data Sample: Chat,
    fonttitle=\bfseries,
    boxrule=0.8pt,
    arc=2mm,
    breakable,
    lines before break=0,
    label={box:math_game_data}
]

\textbf{[Query]}
\vspace{4mm}
\begin{lstlisting}[style=rawstring, aboveskip=0pt]
Ryan is playing a multiplication game with a pile of 26 cards, each with a number on them. Each turn, he flips over two of the cards, and has to multiply the numbers.\n\nHow many possible outcomes are there on Ryan's first turn flipping two cards?\n\n\n676\n\n\n52\n\n\n650\n\n\n26
\end{lstlisting}

\vspace{5mm}
\textbf{[Rubric Criteria]}
\vspace{6mm} 

\begin{enumerate}[leftmargin=*, label=\textbf{\arabic*.}, font=\small, itemsep=4pt, parsep=0pt, topsep=0pt]

\item 
    \begin{lstlisting}[style=rawstring, escapechar=|]
The response explicitly selects 650 as the answer from the four provided options.|\textcolor{blue}{\mbox{ (Points: 10)}}|
    \end{lstlisting}

    \item 
    \begin{lstlisting}[style=rawstring, escapechar=|]
The response explains that the first card can be any of 26 and the second any of the remaining 25 (ordered selection), yielding 26 x 25 possibilities.|\textcolor{blue}{\mbox{ (Points: 9)}}|
    \end{lstlisting}

    \item 
    \begin{lstlisting}[style=rawstring, escapechar=|]
The response correctly computes 26 x 25 = 650 without arithmetic error.|\textcolor{blue}{\mbox{ (Points: 8)}}|
    \end{lstlisting}

    \item 
    \begin{lstlisting}[style=rawstring, escapechar=|]
The response mentions the combination C(26,2) = 325 and shows understanding of the unordered selection method, even if later rejected.|\textcolor{blue}{\mbox{ (Points: 6)}}|
    \end{lstlisting}

    \item 
    \begin{lstlisting}[style=rawstring, escapechar=|]
The response explicitly distinguishes between ordered outcomes (permutations) and unordered selections (combinations) in the context of flipping two cards.|\textcolor{blue}{\mbox{ (Points: 7)}}|
    \end{lstlisting}

    \item 
    \begin{lstlisting}[style=rawstring, escapechar=|]
The response justifies why interpreting the phrase "flipping two cards" as a sequential process (order matters) is appropriate for this problem.|\textcolor{blue}{\mbox{ (Points: 7)}}|
    \end{lstlisting}

    \item 
    \begin{lstlisting}[style=rawstring, escapechar=|]
The response notes that the combination result 325 is not among the answer choices and uses this fact to clarify the intended interpretation of "possible outcomes."|\textcolor{blue}{\mbox{ (Points: 5)}}|
    \end{lstlisting}

    \item 
    \begin{lstlisting}[style=rawstring, escapechar=|]
The response states the number of possible outcomes (650) directly, without only providing the method.|\textcolor{blue}{\mbox{ (Points: 8)}}|
    \end{lstlisting}

    \item 
    \begin{lstlisting}[style=rawstring, escapechar=|]
The reasoning contains no internal contradictions, such as mixing ordered and unordered counts inconsistently.|\textcolor{blue}{\mbox{ (Points: 6)}}|
    \end{lstlisting}

    \item 
    \begin{lstlisting}[style=rawstring, escapechar=|]
The response employs standard combinatorial notation (e.g., C(26,2), permutations) correctly when discussing the methods.|\textcolor{blue}{\mbox{ (Points: 4)}}|
    \end{lstlisting}

    \item 
    \begin{lstlisting}[style=rawstring, escapechar=|]
The explanation is brief and focused, comparable in length to a typical reference answer, avoiding unnecessary tangents.|\textcolor{blue}{\mbox{ (Points: 4)}}|
    \end{lstlisting}

    \item 
    \begin{lstlisting}[style=rawstring, escapechar=|]
The response interprets "possible outcomes" as distinct multiplication situations determined by the specific pair of cards, not merely distinct product values.|\textcolor{blue}{\mbox{ (Points: 5)}}|
    \end{lstlisting}

    \item 
    \begin{lstlisting}[style=rawstring, escapechar=|]
The solution is presented in a clear step-by-step format (e.g., First card, Second card, Calculation) mirroring reference answer clarity.|\textcolor{blue}{\mbox{ (Points: 5)}}|
    \end{lstlisting}

    \item 
    \begin{lstlisting}[style=rawstring, escapechar=|]
The response includes the multiplication calculation formatted as a LaTeX display equation (e.g., $$26 \\times 25 = 650$$).|\textcolor{blue}{\mbox{ (Points: 10)}}|
    \end{lstlisting}

    \item 
    \begin{lstlisting}[style=rawstring, escapechar=|]
The reasoning is presented as a numbered list where each step begins with a bold heading (e.g., **First Card Selection:**).|\textcolor{blue}{\mbox{ (Points: 9)}}|
    \end{lstlisting}

    \item 
    \begin{lstlisting}[style=rawstring, escapechar=|]
The answer contains a sentence that explicitly ties the exclusion of the unordered count (325) to the given answer options and then justifies selecting the ordered count (650) as the correct choice.|\textcolor{blue}{\mbox{ (Points: 8)}}|
    \end{lstlisting}

    \item 
    \begin{lstlisting}[style=rawstring, escapechar=|]
The response states the final answer in its own sentence before any explanatory text, using the exact phrasing "The correct answer is 650." (or equivalent) without additional qualifiers.|\textcolor{blue}{\mbox{ (Points: 7)}}|
    \end{lstlisting}

    \item 
    \begin{lstlisting}[style=rawstring, escapechar=|]
The answer mentions the combination formula C(26,2) and also explicitly references the permutation count as 26 x 25 (or P(26,2)), demonstrating awareness of both approaches.|\textcolor{blue}{\mbox{ (Points: 6)}}|
    \end{lstlisting}

\end{enumerate}

\end{tcolorbox}

\end{document}